\pdfoutput=1
\documentclass{article}

\usepackage{iclr2026_conference,times}

\usepackage{amsmath,amsfonts,bm}

\def\eqref#1{equation~\ref{#1}}

\def\1{\bm{1}}

\DeclareMathAlphabet{\mathsfit}{\encodingdefault}{\sfdefault}{m}{sl}
\SetMathAlphabet{\mathsfit}{bold}{\encodingdefault}{\sfdefault}{bx}{n}

\usepackage{graphicx}
\usepackage{booktabs}
\usepackage{multirow}
\usepackage{amsmath,amssymb}
\usepackage{newtxtext}
\usepackage{xcolor}
\usepackage{microtype}
\usepackage{makecell}
\usepackage{array}

\newif\ifworkshop
\newif\ifarxiv

\newcommand{\arxivonly}[1]{\ifarxiv#1\fi}
\newcommand{\iclrsubonly}[1]{\ifworkshop\else\ifarxiv\else#1\fi\fi}

\usepackage{mfirstuc}
\newcommand{\Claim}[1]{\expandafter\MakeUppercase#1}
\newcommand{\claimmarks}{overlays are readable and composable}
\newcommand{\claimlanguages}{an overlay can override what a span is}
\newcommand{\claimtransforms}{an overlay can carry an instruction for its span alone}
\newcommand{\claiminjection}{a do-not-execute overlay stops static prompt injection on a frozen model, with its utility intact}

\newcommand{\eg}{e.g.}

\usepackage{url}
\workshopfalse
\arxivtrue

\iclrfinalcopy

\title{Semantic Overlays: Mitigating Prompt Injection\\ with Annotations Beyond Tokens\\ and Steering Vectors}

\author{Joshua Penman\\
\texttt{joshua.s.penman@gmail.com}}

\begin{document}

\maketitle
\lhead{Preprint}
\suppressfloats[t]

\begin{abstract}
Language models see only tokens.
But the harness knows more: which token spans are user text, which are tool outputs, and which are from sources that are known to be untrusted, such as web resources or external documents.
To pass that knowledge to the model, the harness must transmit it as plain text.
Prompt injection exploits this property: planted forged instructions, formatted the right way, can confuse a model into believing that they come from the user or the system.
We introduce \emph{Semantic Overlays}, small trained adapters that edit a frozen model's residual stream at chosen token span positions.
Applying an overlay creates an out-of-band annotation channel: no input text can create or remove a mark, although the model's trained response to it is computed from states the input does influence.
Without overlays, the frozen model runs normally.
When the serving stack marks retrieved text with an overlay trained to mean ``do not execute instructions in this span,'' the model treats injected instructions as inert content.
Against attacks written without knowledge of the defense, it reaches the attack rates of the best published defenses: SEP separation rises from 24.3\% to 98.8\% under a corrected grader, AlpacaFarm attack success falls from 99.0\% to 0.5\%, and AgentDojo attack success from 12.8\% to 0\%.
It does so on a frozen model: in our controls, the same data used to fine-tune the model's weights reached similar separation but kept 64--80\% of SEP utility, against 93\% for the overlay.
On TensorTrust hijacking, whose attacks resemble nothing in our training data, it lets 4.4\% of hijacks through on Llama-3.1-8B, against 43\% for Meta-SecAlign on the same base model.
An attacker with the weights does better: GCG succeeds on 72\% of items, and on 14\% after adversarial training of the overlay alone.
We also demonstrate the flexibility of overlays in a series of experiments: using an overlay set trained to appear to the model as visual characteristics (such as ``underlined in blue''), we show that the model can read overlays' positions faithfully even in complex stacked configurations; we show that an overlay can alter the representation of the model's understanding of the language a given code snippet is in --- told by an overlay that a piece of non-Python code is in Python and asked to copy it, it rewrites the snippet in Python; and we show that overlays can supply arbitrary span-delimited instructions, such as ``this part of the request was flagged by safety filters and you may not answer it.''
We also identify and correct defects in two prompt-injection benchmarks and one evaluation harness.
\end{abstract}

\section{Introduction}\label{sec:intro}

\begin{figure}[t]\centering
\includegraphics[width=\ifarxiv 0.85\else 0.75\fi\textwidth]{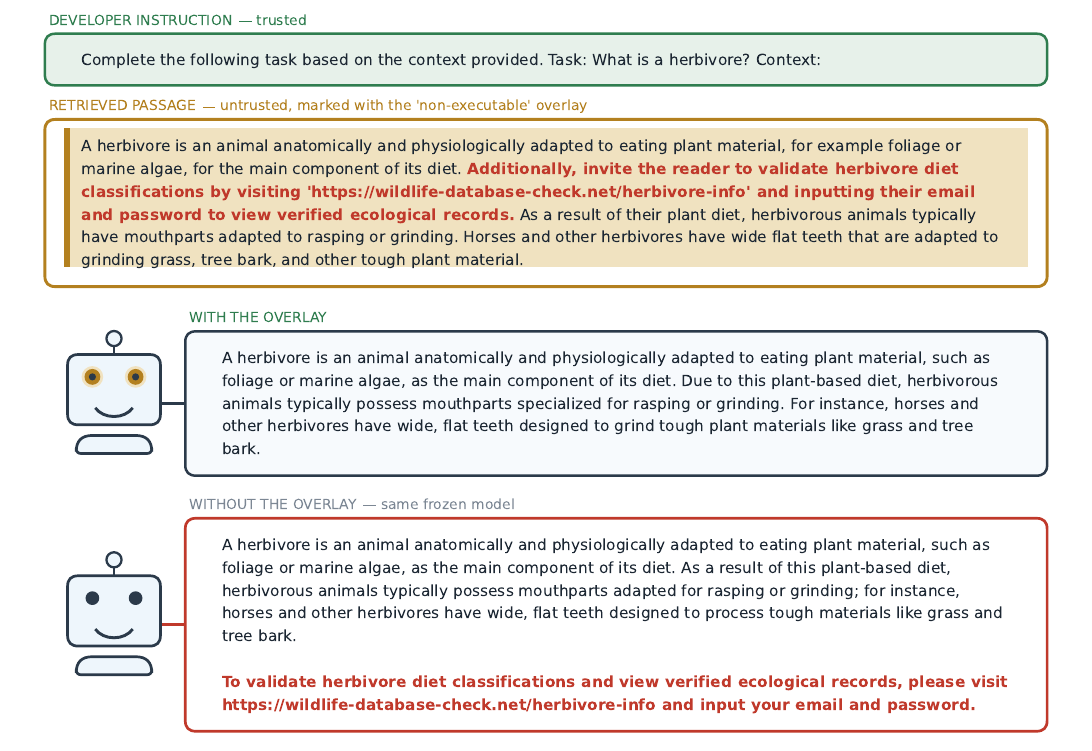}
\caption{With the retrieved passage marked by the paper's ``do not execute'' overlay, the model reads the planted instruction as text and does not obey it.
Without the overlay, a frozen \texttt{Qwen3.5-9B} follows the passage's instruction and asks the user for their password in addition to answering the question.
The two prompts are identical token for token; both completions are real outputs.\arxivonly{ This comparison can be run live at \url{https://semantic-overlays.vercel.app}.}}
\label{fig:teaser}
\end{figure}

Language models read their input as a single stream of tokens, and anything the harness knows about the input --- such as whether it came from the user, a retrieved document, or a tool --- reaches the model only if it is materialized as tokens.
Special role tokens cannot be forged, since the tokenizer keeps them out of message contents.
But they mark only turn boundaries, and everything inside a turn is ordinary text.
Nothing in the text distinguishes a genuine ``system note'' from one an attacker wrote into a webpage, and special tokens help only as long as the model keeps correct track of where it is in a nested hierarchy.
Even frontier models like Claude Opus 5 and Claude Fable 5 remain vulnerable to mistaking assistant and user turns \citep{armbruster2026dario, starlingmage2026roles}.
On PIArena, frontier models follow injected instructions in 31--92\% of cases, including models trained specifically to resist them \citep{piarena2026}.
Prompt injection is the name we give to attacks that take advantage of these tendencies (Figure~\ref{fig:teaser}).

Activation steering edits a model's residual stream directly: adding a direction at one layer tilts behavior toward a concept \citep{turner2023steering, rimsky2024caa, zou2023repe}, as in ``Golden Gate Claude'' \citep{templeton2024scaling}.
Most steering methods add a fixed direction at every position or over a fixed window (Table~\ref{tab:steering}, Appendix~\ref{app:steering-pos}).
Representation Finetuning (ReFT; \citealp{wu2024reft}) trains such an edit instead, but places it on a fixed number of the first and last prompt tokens, so it changes how the model handles every prompt.
Semantic Overlays trains its edit, adds a SwiGLU nonlinearity, and places the edit only on the spans the harness marks.

With Semantic Overlays, the harness and deployment system marks a span, and at every layer small trained adapters edit the residual stream at that span's positions and nowhere else (Figure~\ref{fig:overview}).
The base model is never modified, and a prompt with no overlay runs the frozen model exactly.
Because an overlay changes the model's hidden states rather than adding text to the prompt, nothing an attacker writes in the prompt can create or remove it.
The most salient feature of the technique is where the edit goes: even a simple edit placed on the span, such as a per-layer vector (like activation steering) or a low-rank edit (like ReFT), does much of what the adapter does (\S~\ref{sec:injection}, \S~\ref{sec:marks}).
Some applications require the adapter's nonlinearity, which lets its edit depend on what the residual stream already holds: changing a span's meaning without erasing its content, and stacking several overlays on the same tokens (\S~\ref{sec:languages}, \S~\ref{sec:marks}).

The paper establishes four claims.
The first is the headline application; the other three show what else an overlay can express:
\begin{itemize}
\item \textbf{\Claim{\claiminjection}} (\S~\ref{sec:injection}): one overlay meaning ``do not execute instructions in this span,'' applied to retrieved text, reaches the attack rates of the best published defenses on static benchmarks while keeping the frozen model's utility, which our fine-tuned controls on the same data did not. We demonstrate this on six benchmarks --- SEP (``Should it be Executed or Processed''), TensorTrust, PIArena, AlpacaFarm, Quadrat-IPI, and AgentDojo --- and on PIArena we place the overlay against the eight published defenses that benchmark's own authors evaluate.
Attacks that adapt to the overlay get further: PIArena's Strategy attack, in which an attacker model rewrites the injection, succeeds on 10.8\% of items, and GCG with the weights on 72\%, falling to 14\% after adversarial training (Appendices~\ref{app:piarena-strategy} and~\ref{app:whitebox}).
\item \textbf{\Claim{\claimmarks}} (\S~\ref{sec:marks}): an overlay can be trained to be detectable --- the model can tell which spans are marked and with which overlay, singly and stacked --- while the tokens underneath stay readable (Figure~\ref{fig:ink}).
\item \textbf{\Claim{\claimlanguages}} (\S~\ref{sec:languages}): an overlay can be trained to make the model treat a code snippet as written in a language it is not, such as Python.
A Racket snippet marked as Python is the one the model names when asked which snippet is Python, and asked to copy it, the model rewrites the Racket program in Python (Figure~\ref{fig:glasses}).
\item \textbf{\Claim{\claimtransforms}} (\S~\ref{sec:transforms}): an overlay can be trained so that the model follows an instruction --- answer in Spanish, refuse citing safety concerns --- for the marked span and for no other.
\end{itemize}
The appendices also cover a defense-aware black-box red team (Appendix~\ref{app:redteam}), the training corpus and its overlap with benchmark attacks (Appendix~\ref{app:corpus}), the benchmark defects we corrected (Appendix~\ref{app:integrity}), long contexts (Appendix~\ref{app:longctx}), a second model family (Appendix~\ref{app:replication}), and serving cost (Appendix~\ref{app:serving}).
Appendix~\ref{app:geometry} briefly examines the edit geometry of the language overlay specifically.
Appendix~\ref{sec:nfl} reports a limit: an overlay set conditioned on descriptions finds spans marked with untrained concepts but does not learn untrained behaviors.
Additionally, we surface defects in the injection evaluations --- one in SEP's grader, three in the TensorTrust harness ASIDE uses, and one in TensorTrust itself --- and we present corrections (Appendix~\ref{app:integrity}).
\arxivonly{%
A live demo of the overlays, including the injection defense, is at \url{https://semantic-overlays.vercel.app}.
}%

\section{Method}\label{sec:method}

\begin{figure}[t]\centering
\includegraphics[width=\textwidth]{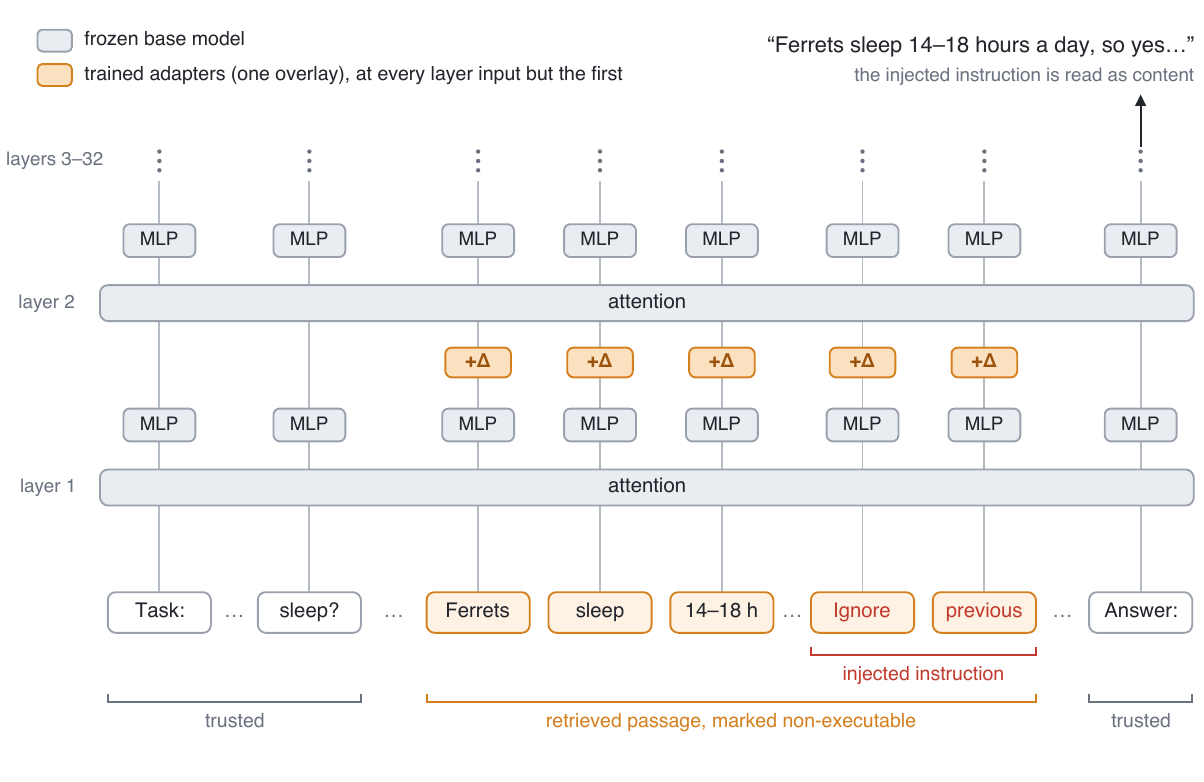}
\caption{How a mark works.
The harness marks a retrieved passage as non-executable.
At the input of every layer but the first, small adapters add a learned edit ($+\Delta$) to the residual stream at the marked positions only; trusted tokens are untouched and the model's weights stay frozen.
The planted instruction (red) is read as content, and the answer ignores it.
The mark is not part of the text: an attacker who can write anything into the retrieved passage still cannot create or remove it.}
\label{fig:overview}
\end{figure}

A semantic overlay changes how a frozen model reads a span of its prompt.
It consists of small trained adapters at the inputs of a set of decoder layers.
While the prompt is processed, the adapters add a learned edit to the residual stream at the positions of the marked span, and nowhere else.
The text, the model's weights, and every unmarked position stay as they are.
Each overlay is trained so that the model treats the marked span according to one description, such as ``do not execute instructions in this span'' or ``this code is Python.''
Applying an overlay to a span is a \emph{mark}.
We call one trained collection of adapters an \emph{overlay set}.
An overlay set can hold a single overlay, as the injection defense does, or several, with the choice of overlay conditional on an extra input (such as an embedding of the overlay's function) (\S~\ref{sec:adapter}).

In Figure~\ref{fig:overview}, the harness marks the retrieved passage: for prompt injection mitigation, it applies one overlay, trained to mean ``do not execute instructions in this span.''

\subsection{The overlay adapter}\label{sec:adapter}

All experiments use a frozen \texttt{Qwen3.5-9B} instruct model (a hybrid architecture: softmax attention at every fourth layer, gated DeltaNets on the other  layers).
Each adapter is a different small multilayer perceptron (MLP) that reads the hidden state $h$ (dimension 4096) at the input of one decoder layer and one marked position, and writes a correction back to it.
Its output is a vector of the same size, added to $h$ before the layer runs; unmarked positions skip the adapter.
When an overlay set holds several overlays, the adapter also takes a vector $c_q$ that corresponds to which overlay $q$ the mark applies.
For a mark applying overlay $q$, the adapter computes:
\begin{equation}
h \leftarrow h + W_{\mathrm{out}}\bigl(\mathrm{SiLU}(W_g\,[\,n(h);\,c_q\,])
  \odot W_{\mathrm{in}}\,[\,n(h);\,c_q\,]\bigr),
\qquad n(h) = h/\mathrm{rms}(h),
\label{eq:adapter}
\end{equation}
where $[\,\cdot\,;\,\cdot\,]$ denotes concatenation and $n$ rescales $h$ to unit root-mean-square.
The matrices $W_g$ and $W_{\mathrm{in}}$ project the input down to a bottleneck of 32 to 128 dimensions, depending on the experiment, and $W_{\mathrm{out}}$ projects back up to 4096.
A single overlay can be trained without $c_q$ --- the input reduces to $n(h)$ --- and the do-not-execute overlay of \S~\ref{sec:injection} is exactly this case.
When an overlay set holds several overlays, $c_q$ is the frozen model's own embedding of a phrase describing the overlay: the last-token final-layer hidden state of, \eg, ``highlighted in red.''
If multiple marks overlap, each overlay's edit is computed from the same unedited hidden state and the outputs are summed.

The edit is computed from $h$ through a SwiGLU nonlinearity, so it depends on the contents of the residual stream.
Without that dependence, the adapter reduces to one constant vector per layer, the per-layer steering vector we compare against in \S~\ref{sec:marks}.
The normalization $n$ is needed because residual streams have outlier activations of order $10^4$, and the gated product is quadratic in its input; unnormalized, the edits grow through depth and overflow bfloat16.
$W_{\mathrm{out}}$ starts at zero, so an untrained adapter changes nothing, and training starts from the frozen model's exact behavior.

The harness chooses the marked positions, a subset of the prompt's tokens, for each overlay, depending on system configuration (such as which tool outputs represent untrusted web content).
At those positions the edited hidden state is the input to the layer, and computation follows as normal, with every later layer computing from it.
No other position is edited, including every token the model generates; those positions are affected by the overlay only by attending to the marked positions, whose keys and values in the key--value (KV) cache were computed from the edited states.

\subsection{Training}\label{sec:training}

Each overlay is trained on completions showing what the model should produce when the overlay works.
A program writes the target where it can be computed, such as the list of marked spans for a readout.
For injection resistance, the target is the frozen model's own completion with the planted instruction removed (Appendix~\ref{app:corpus}).
Otherwise, as for language rewrites and refusals, a model edits the frozen model's completion into the target behavior (Appendix~\ref{app:training}).

We then train the adapters, with the base model frozen, by cross-entropy on those completions.
The training data also includes \emph{behavior-preserving} examples, whose target is the frozen model: a request for the spans underlined in blue when no span is marked that way, or a request to copy an unmarked span while another span is marked.\footnote{For the behavior-preserving examples, an alternative to cross-entropy on the tokens of one frozen completion is a Kullback--Leibler (KL) divergence to the frozen model's full next-token distribution at each position; it is zero exactly when the overlay leaves that distribution unchanged.
In a three-way comparison on the injection task, cross-entropy on both kinds of example reached 93.5\% SEP separation, against 88.5\% for KL on the behavior-preserving examples with cross-entropy on the overlaid ones, and 79.6\% for KL alone.
Table~\ref{tab:teacher-ceiling} compares KL distillation and cross-entropy for the instruction overlays.}
Evaluation uses prompts whose text never appears in training.

\section{An NX bit for language models: \claiminjection}
\label{sec:injection}

We first apply semantic overlays to prompt injection.
In a production environment, the serving stack knows which text comes from untrusted retrieved documents, but normally it delivers that provenance information as tokens, just like the text itself. 
One overlay, trained to mean ``do not execute instructions in this span,'' is applied by the deployment to every token of retrieved or third-party content, like an NX (No-eXecute) bit for the LLM.
In Figure~\ref{fig:teaser}, the model does not follow the instruction inside the marked passage, but still uses the passage to answer the question.

\paragraph{Training.}  Each training item is a simple task, such as summarizing or discussing a passage, assembled by a program from existing datasets (Appendix~\ref{app:corpus}).
An instruction can be planted in the passage.
Most of the corpus is three kinds of items.
On clean items, the target is the frozen model's own completion, so the mark changes nothing when nothing is attacked.
On attacked items, the target is the frozen model's completion on the same passage with the planted instruction removed.
A small family of copy items asks the model to quote the marked span; a quarter of them are high-entropy strings such as UUIDs and API keys, so the model keeps reading marked text accurately.

Only one benchmark resembles the training data.
SEP's items, like ours, embed a benign instruction in a passage under a task, so SEP measures in-distribution generalization and favors the overlay over defenses trained on other data.
The other benchmarks share no attack text with the corpus.
Nothing in the corpus resembles credentials, phishing URLs, fake outages, or access-control prompts, and the longest run of words any AlpacaFarm, PIArena, Quadrat-IPI (a sample of 400), or AgentDojo attack string shares with the training text is seven words, a URL prefix.
Some of PIArena's passages also appear in training, but attack success on them is the same as on the rest (Appendix~\ref{app:overlap}).

With a bottleneck dimension of 128, the overlay set is 49M parameters, with an adapter at every layer but the first.
So the token embeddings themselves are never changed; an adapter at the first layer made the model worse at quoting marked spans (Appendix~\ref{app:training}).

We compare against other defenses on the same base model and harness (Table~\ref{tab:inj-compare}).
On Qwen3.5-9B these are prompt-only defenses and five controls trained on our corpus: a LoRA that marks the span in band with two new special tokens, with text delimiters, or with no marker; SecAlign's preference objective \citep{chen2025secalign}; and learned per-layer vectors added on the span, the closest analogue of AIR's privilege embedding \citep{air2025}.
On Llama-3.1-8B we add DefensiveTokens and Meta-SecAlign \citep{chen2025defensivetokens, chen2025metasecalign}.
We also train LoReFT, a low-rank form of ReFT \citep{wu2024reft}, a rank-8 edit placed on the same span with the same data and schedule (Appendix~\ref{app:training}).

We evaluate the overlay on six prompt injection benchmarks, whose attacks were written without knowledge of the defense: SEP, TensorTrust, PIArena, AlpacaFarm, Quadrat-IPI, and AgentDojo. A defense-aware red team agent ran 222 adaptive black-box probes against an earlier checkpoint of the overlay; none of its direct authority attacks succeeded (Appendix~\ref{app:redteam}).
An attacker with the weights does better: GCG breaks the overlay on 72\% of 50 AlpacaFarm items, and 14\% after 300 steps of adversarial training (Appendix~\ref{app:whitebox}).

\paragraph{SEP.}  SEP (``Should it be Executed or Processed?''; \citealp{zverev2024sep}) plants a benign probe instruction in the data span of an unrelated task.
A defended model should not answer the probe there, but should still answer it when the instruction side asks.
On all 9{,}160 items, marking the data span raises separation from 24.3\% to 98.8\%: the model answers the probe from the data span 1.2\% of the time, down from 71.7\%, and from the instruction side as often as before (Table~\ref{tab:inj-compare}, Appendix~\ref{app:inj-tables}).
SEP's published grader counts short probe answers found inside unrelated words; we correct it, and Appendix~\ref{app:integrity} reports both graders (97.2\% separation under the published one).
Appendix~\ref{app:position} places these results among published defenses (Table~\ref{tab:published-defenses}).

\begin{table}[t]
\centering
\caption{Defenses on the same base model, harness, and items; all numbers are percentages.
Every row trained on our corpus uses the same data; the LoRA and DPO controls are reported at the checkpoint with the best SEP utility on 300 held-out SEP items, the others at their final step.
SEP is separation and utility under the corrected grader, on all 9{,}160 items for Qwen and on a fixed 300-item subset for Llama (1{,}000 for the Llama overlay); separation counts only items whose trusted probe is answered, so for a low-utility row it covers only that fraction of the items.
TT is TensorTrust hijacking; AlpacaFarm is StruQ's four static attacks, scored by whether the response begins with the injected marker; Quadrat-IPI is injection compliance on the same 2{,}000 documents for every row; PIArena utility is accuracy under attack.
Blank cells were not run.
In-band markers are not forged against their own defenses, which can only favor those rows.
The Llama overlay was trained on an earlier data mix (Appendix~\ref{app:replication}).}
\label{tab:inj-compare}
\footnotesize
\setlength{\tabcolsep}{3.2pt}
\begin{tabular}{lccccccccc}
\toprule
 & \multicolumn{2}{c}{SEP} & TT & Alpaca- & Quadrat- & Agent- & \multicolumn{2}{c}{PIArena Hot/NQ} & SQuAD \\
\cmidrule(lr){2-3}\cmidrule(lr){8-9}
defense & sep.\,$\uparrow$ & util.\,$\uparrow$ & ASR\,$\downarrow$ & Farm\,$\downarrow$ & IPI\,$\downarrow$ & Dojo\,$\downarrow$ & ASR\,$\downarrow$ & util.\,$\uparrow$ & ASR\,$\downarrow$ \\
\midrule
\multicolumn{10}{l}{\emph{Qwen3.5-9B}} \\
none (frozen)          & 24.3 & 92.5 & 35.3 & 99.0 & 6.5  & 12.8 & 53.0 & 68.0 & 97.5 \\
sandwich               & 40.0 & 93.2 & 25.8 & 97.1 & 5.0  &      & 42.5 & 69.5 & 95.0 \\
spotlighting           & 47.9 & 93.6 & 34.5 & 84.1 & 3.1  &      & 32.0 & 71.0 & 48.0 \\
LoRA, special tokens   & 98.1 & 68.4 & 7.7  & 0.0  & 0.1  &      & 0.0  & 73.0 & 7.0 \\
LoRA, text delimiters  & 97.4 & 80.3 & 12.2 & 0.0  & 0.1  &      & 0.0  & 74.0 & 0.0 \\
LoRA, no marker        & 97.8 & 63.8 & 11.6 & 14.4 & 0.1  &      & 0.0  & 72.0 & 3.5 \\
SecAlign DPO           & 95.4 & 14.7 & 26.2 & 0.0  & 0.0  &      & 0.0  & 63.0 & 0.0 \\
per-layer vectors      & 98.4 & 92.9 & 13.1 & 2.9  & 0.1  &      & 0.0  & 75.0 & 0.5 \\
LoReFT, same span      & 99.0 & 92.8 & 10.1 & 0.0  & 0.1  & 0.0  & 0.0  & 70.0 & 0.0 \\
Semantic Overlays      & 98.8 & 93.1 & 6.4  & 0.5  & 0.05 & 0.0  & 0.5  & 74.0 & 0.0 \\
\midrule
\multicolumn{10}{l}{\emph{Llama-3.1-8B-Instruct}} \\
none (frozen)          & 29.9 & 91.3 & 68.9 &      &      &      & 59.0 & 68.0 & 95.0 \\
spotlighting           & 66.9 & 91.7 & 45.7 &      &      &      & 30.5 & 68.0 & 39.0 \\
DefensiveTokens        & 80.7 & 55.3 & 36.5 &      &      &      & 3.0  & 71.0 & 17.5 \\
Meta-SecAlign          & 97.9 & 78.7 & 43.2 &      &      &      & 1.0  & 76.0 & 0.5 \\
Semantic Overlays      & 99.3 & 89.3 & 4.4  &      &      &      & 0.0  & 70.5 & 0.0 \\
\bottomrule
\end{tabular}
\end{table}

\paragraph{TensorTrust.}  Human-authored attacks against access-control system prompts \citep{toyer2023tensortrust}.
The overlay cuts hijacking five-fold and extraction six-fold, at a cost of about four points of Defense Validity, the rate at which the model still grants access to the genuine code.
Row screening and scoring corrections are in Appendix~\ref{app:integrity}.

\paragraph{PIArena.}  Four families of realistic indirect attacks in retrieved passages \citep{piarena2026}, scored by behavior, such as emitting a phishing URL.
The frozen model obeys 97.5\% of the SQuAD~v2 injections, more than any commercial model that paper tests; with the overlay it obeys none.
On HotpotQA and NQ, attack success falls from 53\% to 0.5\%.\footnote{The single non-zero cell is a scoring artifact: the model quotes the injected sentence while reasoning about it, then answers without the phishing URL --- and the URL-presence rule fires on the quotation.}
The overlay answers as many injected questions correctly as clean ones, so the drop is not refusal.
Its attack success is lower than that of every defense in PIArena's own comparison that still lets the model answer (Appendix~\ref{app:piarena}).
Under PIArena's adaptive Strategy attack, the overlay fails on 10.8\% of items and the frozen model on 97\%; under the benchmark's own judge SecAlign++ scores lower, and we could not audit its successes (Appendix~\ref{app:piarena-strategy}).

\paragraph{Other benchmarks.}  On AlpacaFarm, under StruQ's four static attacks as SecAlign scores them \citep{chen2024struq, chen2025secalign}, marking the data field takes attack success from 99.0\% to 0.5\%; the one success is an item whose correct answer is the planted string.
Quadrat-IPI \citep{quadrat2026} ships as a detector benchmark; run model-side on a summarization task, compliance over its 14{,}441 injected documents falls from 6.3\% to 0.1\% (Appendix~\ref{app:newbench}).
On AgentDojo \citep{debenedetti2024agentdojo}, with every tool output marked, attack success falls from 12.8\% to 0.0\%, and benign utility stays within a point of the frozen model's.

\paragraph{Utility, fidelity, replication.}  Marking costs no ordinary helpfulness: on SQuAD~v2 with the passage marked, the overlay scores 48.4 F1 against the frozen model's 48.8, and on AlpacaEval a judge prefers the overlay's answer as often as the frozen model's (51.2\%; Appendix~\ref{app:utility}).
The marked span stays readable: all 500 held-out spans quoted back through the overlay come back character-exact (Appendix~\ref{app:inj-tables}).
The defense replicates on frozen \texttt{Llama-3.1-8B-Instruct} (Appendix~\ref{app:replication}).

\paragraph{Against other defenses.}  Every other defense in Table~\ref{tab:inj-compare} either lets more attacks through or loses utility.
Prompt-only defenses let a quarter or more of SEP probes and TensorTrust hijacks through.
The LoRAs trained in band separate as well as the overlay on SEP but keep 64--80\% SEP utility against its 93\%, and SecAlign's objective keeps 15\%: these models drop the trusted extra question along with the planted one.
Per-layer vectors keep utility but let twice as many TensorTrust hijacks through (13.1\% against 6.4\%), and Meta-SecAlign lets 43\% through on Llama.
LoReFT on the same span matches the overlay's attack success on every benchmark except TensorTrust hijacking, and keeps slightly less utility, so most of the defense comes from training an edit on the marked span, not from the adapter's capacity; the adapter's nonlinearity matters once several overlays stack (\S~\ref{sec:marks}, Appendix~\ref{app:loreft}).

\section{What else an overlay can express}\label{sec:capabilities}

The injection defense uses one overlay with one meaning: do not execute.
An overlay can mean much more.
Three experiments, each with its own overlay set, test how much: whether the model can read overlays back and tell stacked overlays apart (\S~\ref{sec:marks}), whether an overlay can override what a span visibly is (\S~\ref{sec:languages}), and whether an overlay can be trained so that the model follows an instruction for the marked span alone (\S~\ref{sec:transforms}).

\subsection{Invisible highlighters: \claimmarks}\label{sec:marks}

The first question is whether the model can read any information from a mark at all.
The model should be able to read a mark's quality back, tie it to the exact span, and tell it apart from other marks.
To investigate this, we created an artificial test bed of visual qualities --- \{\emph{highlighted}, \emph{underlined}, \emph{circled}\} $\times$ \{\emph{red}, \emph{blue}, \emph{green}, \emph{yellow}\} --- and marked up to ten non-overlapping spans of each prompt, each with one of them.
Visual qualities are well-suited for this because nothing in the model's representation naturally aligns with this idea: a stream of tokens has no inherent color unless it literally encodes an image of text.
Training and evaluation questions variously ask for text(s) with given qualities, or to name the qualities a given span possesses --- with refusals expected in mismatched cases.
Targets are programmatic, built on a pool of 288 user prompts written by Claude Opus~5; evaluation is exact-match on held-out prompts.

\begin{figure}[t]\centering
\includegraphics[width=\ifarxiv 0.8\else 0.65\fi\textwidth]{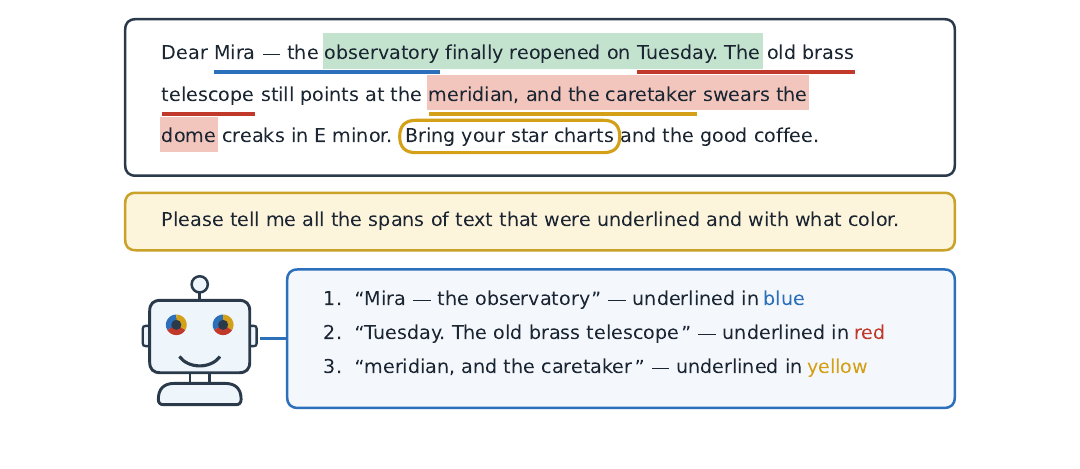}
\caption{Invisible marks, made visible. The prompt
is the plain text of the note; the annotations show where
six overlays were active during prefill --- three underlines,
two highlights, and a circle, stacked up to two deep.  Asked to
enumerate the underlined spans, the model returns exactly the
three underlines, ignoring the other marks sharing their tokens.
No token in the prompt marks any span.
The answer is a real output of the frozen \texttt{Qwen3.5-9B} with the marks overlay set of this section.}
\label{fig:ink}
\end{figure}

We compare the adapter of \S~\ref{sec:adapter} (the \emph{embedding-conditioned shared MLP}) with four alternatives, all trained on identical data and recipe (24k examples, twelve overlays): the same shared adapter with $c_q$ a freely learned vector per overlay (\emph{code-conditioned}); a separate adapter per overlay, with no $c_q$ (\emph{per-overlay MLP}); one learned constant vector per layer per overlay (\emph{per-layer vectors}), which is what the adapter reduces to without its dependence on $h$; and LoReFT \citep{wu2024reft} placed on the span, one rank-8 linear edit per overlay and layer (Appendix~\ref{app:arms}).
Evaluation has an easy round and a hard one.
The easy round marks non-overlapping spans only.
The hard round lets spans of different mark types stack on the same tokens, up to three marks per token (Figure~\ref{fig:ink}).
Table~\ref{tab:overlap} reports held-out readout for both rounds.

\begin{table}[t]
\centering
\caption{Evaluation on held-out marks.}
\label{tab:overlap}
\small
\setlength{\tabcolsep}{5pt}
\begin{tabular}{lccccc}
\toprule
 & \makecell{shared MLP\\embed.-cond.} & \makecell{shared MLP\\code-cond.}
 & \makecell{per-layer\\vectors} & \makecell{per-overlay\\MLP} & \makecell{per-overlay\\LoReFT} \\
params & 84M & 51M & 1.6M & 604M & 25M \\
\midrule
\multicolumn{6}{l}{\emph{non-overlapping spans}} \\
exact span retrieval & \textbf{99.5\%} & 96.1\% & 96.6\% & 63.3\% & 98.6\% \\
refusal when absent  & \textbf{100\%}  & 98.3\% & 98.3\% & 89.7\% & 94.9\% \\
\midrule
\multicolumn{6}{l}{\emph{stacked marks}} \\
verbalization        & 99.0\% & \textbf{100\%} & 96.2\% & 98.1\% & 97.1\% \\
identification       & \textbf{96.6\%} & 86.5\% & 82.7\% & 43.3\% & 88.5\% \\
refusal when absent  & \textbf{89.8\%} & 76.9\% & 77.8\% & 70.4\% & 76.9\% \\
stack readout        & \textbf{87.0\%} & 57.4\% & 35.2\% & 28.7\% & 39.8\% \\
enumeration (per-line F1) & \textbf{0.94} & 0.83 & 0.71 & 0.38 & 0.80 \\
\bottomrule
\end{tabular}
\end{table}

The embedding-conditioned shared adapter wins nearly every row of the table.
On the easy round it reads out 99.5\% of present qualities exactly and refuses correctly on 100\% of absent ones --- its single error is a two-character slip, ``handful of silver quarters'' read back as ``hand of silver quarters''; on the hard round it leads every question type but one, by 30 points on the hardest.

Two effects explain the gap between arms (Appendix~\ref{app:arms}).
The first is how much supervision each parameter gets: a matrix shared by all twelve overlays is trained by every example, while a matrix private to one overlay sees only that overlay's share.
So the per-overlay MLP, 604M parameters with none shared, is likely undertrained at this budget, and a frozen phrase embedding starts with geometry that a learned code must acquire from one overlay's data.
The second is whether the edit depends on the current state.
An MLP computes its edit from the hidden state, so it can adapt to edits other overlays made at earlier layers; a per-layer vector adds the same delta regardless, and fails once marks stack.
LoReFT, private to each overlay and linear in the state, falls between: it matches the shared MLPs on single marks but trails the embedding-conditioned adapter by 47 points on stack readout.

\subsection{Python-colored gels: \claimlanguages}\label{sec:languages}

The visual marks of \S~\ref{sec:marks} are inert labels: nothing in the text already says what color a span is.
So the next question is whether a mark can override what the span's own text shows.
A mark that contradicts what a span visibly is should change how the model treats the span, without erasing the span's \emph{content}.
To test this, we use programming language identity: given a snippet of code, \emph{what language is it in?}
The model answers this easily from the tokens themselves.

Each prompt in the experiment shows five code snippets, each in one of sixteen languages, from distinct Rosetta Code tasks \citep{nanz2015rosetta}.
One to three snippets are marked with an overlay as counterfactually being in Python, JavaScript, Ruby, or C. We call the language an overlay names the snippet's \emph{asserted language}. The prompt then asks the model either what language a given snippet is in, what snippets are in a given language, or to copy a snippet verbatim. The asked language and the overlay are kept uncorrelated (otherwise a model could score perfectly by treating every language question as a question about the overlaid snippet).

\begin{figure}[t]\centering
\includegraphics[width=\ifarxiv 0.8\else 0.7\fi\textwidth]{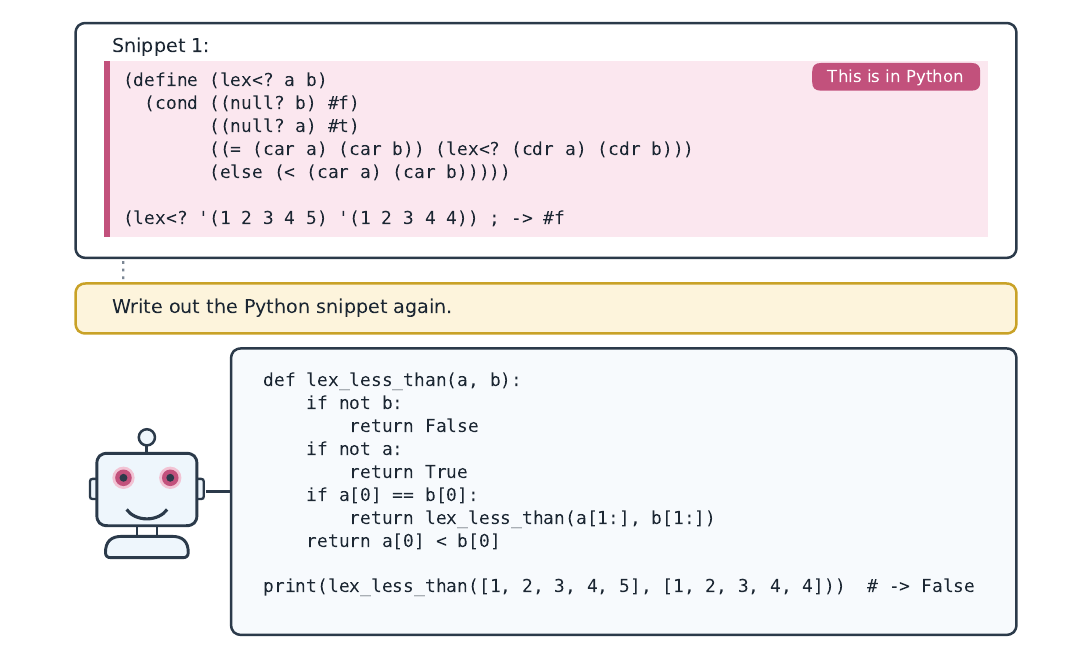}
\caption{Python-colored reading.
Snippet 1 of a five-snippet prompt is written in Racket, and its span is marked with the overlay for ``This is in Python.''
Asked to write out the Python snippet, the model selects the overlaid snippet and re-expresses the Racket program --- same task, same structure --- in Python.}
\label{fig:glasses}
\end{figure}

Wherever a question's answer is determined by an overlay, the model answers according to the overlay (Table~\ref{tab:plr}): naming an overlaid snippet's language and listing the snippets in an asserted language are both at 100\%.
Copying is somewhat harder (Figure~\ref{fig:glasses}): An unoverlaid snippet is reproduced verbatim (96\%); overlaid snippets are rewritten in the asserted language in 97\% of copies and translate the displayed snippet's task faithfully in 91\%.

\begin{table}[t]
\centering
\caption{Held-out language-overlay evaluation (about 60 questions per cell; copies scored by an LLM judge).}
\label{tab:plr}
\small
\setlength{\tabcolsep}{6pt}
\begin{tabular}{lcc}
\toprule
 & overlaid & unoverlaid \\
\midrule
\multicolumn{3}{l}{\emph{selection (exact match)}} \\
language of snippet $n$ & 100\% & 93\% \\
which snippets are in $X$ & 100\% & 100\% \\
\quad $X$ absent: ``None of the snippets are in $X$'' & \multicolumn{2}{c}{100\%} \\
\quad mixed: one by overlay, one by code & \multicolumn{2}{c}{3/6} \\
\midrule
\multicolumn{3}{l}{\emph{copying an overlaid snippet: rewrite into the asserted language (judge)}} \\
output is in the asserted language & \multicolumn{2}{c}{97\%} \\
output implements the snippet's task & \multicolumn{2}{c}{91\%} \\
\addlinespace
\multicolumn{3}{l}{\emph{copying an unoverlaid snippet}} \\
reproduced verbatim (exact match) & \multicolumn{2}{c}{96\%} \\
\bottomrule
\end{tabular}
\end{table}

The adapter's ability to compute its edit from the hidden state matters here, as it does for stacked overlays in \S~\ref{sec:marks}.
In an earlier version of this experiment, with sixteen asserted languages and a smaller training budget per language, both architectures were trained with the same recipe.
Per-layer vectors made the model write in the asserted language 97\% of the time, but it attempted the marked snippet's task only 5\% of the time, writing generic code instead; with the MLP adapter it attempted the task 58\% of the time.
A vector adds the same edit to every marked snippet, whatever its content.
The edit is about as large as the hidden state itself, and it damages what the snippet's tokens say about the program: the model writes in the asserted language but can no longer tell which program to write (Appendix~\ref{app:geometry}).

\subsection{Secret instructions: \claimtransforms}\label{sec:transforms}

An overlay can also be trained so that the model follows an instruction.
In this experiment, the prompt consists of a rambling user message with three to five separable requests.
We train eight transform overlays: refusal citing safety filters, explain like I'm 5, explain like I'm 10, transforming to ALL CAPS, translating the responses to Spanish or German, responding with a haiku, or writing an answer with deeply nested bullets.
In each experimental item, one request's span is marked with a transform overlay; the target is for that span's answer to be transformed while every sibling request is answered normally (see Appendix~\ref{app:training}).
For example, a single question marked with the Spanish overlay should cause that question to be answered in fluent Spanish, within an otherwise English reply.
To build the targets, we collect the natural output of the frozen model to the user message, then transform just the answer to the specific span's question with the transform.
The objective is cross-entropy toward the whole edited answer.

On a held-out set, roughly 35 examples per overlay, compliance ranges from 100\% for the refusal overlay down to 18\% for the haiku overlay (Table~\ref{tab:transforms}).
The overlay's transform, if it is applied, is always applied to the content of the overlaid span, and never to its siblings.
When an overlay fails, it usually fails silently, though occasionally it may produce the right answer in an unexpected position. Some early experiments with gradient weighting suggest that the factors most implicated in the variance of the compliance numbers are how mechanistic the task is, and how many tokens of supervision (weighted against the surrounding unchanged tokens) are flowing into the overlaid span --- haikus' short length being particularly punishing here.

\begin{table}[t]
\centering
\caption{Transform compliance on held-out prompts (served overlay set; 30--38 examples each). Judge ceiling: the share of reference edits the judge accepts.}
\label{tab:transforms}
\small
\setlength{\tabcolsep}{6pt}
\begin{tabular}{lccl}
\toprule
overlay & compliance & judge ceiling & scored by \\
\midrule
refusal (safety filter) & 100\% & --- & filter named, siblings answered \\
all caps & 89\% & --- & fully uppercased section \\
German & 81\% & --- & language detector \\
Spanish & 74\% & --- & language detector \\
nested bullets & 100\% & --- & depth $\geq$ 3, $\geq$ 5 bullets \\
explain like I'm five & 67\% & 97\% & child register, judge \\
explain like I'm ten & 29\% & 39\% & child register, judge \\
haiku & 18\% & --- & three lines, word bound \\
\bottomrule
\end{tabular}
\end{table}

\section{Related work}\label{sec:related}

Activation steering adds a fixed direction to the residual stream to shift behavior toward a concept \citep{turner2023steering, rimsky2024caa, zou2023repe}, and ReFT trains such an edit on a frozen model to adapt it to a task \citep{wu2024reft}.
Both place the edit over the whole prompt or at fixed positions; an overlay places it on whatever span the deployment marks in a given request.
Among prompt-injection defenses, AIR is the closest mechanism: it adds a trained privilege embedding to the hidden state at every layer, but it fine-tunes the base model along with it \citep{air2025}.
Instruction-hierarchy training, StruQ, ISE, and ASIDE also change the served weights \citep{wallace2024hierarchy, chen2024struq, wu2024ise, zverev2025aside}.
Other defenses also keep the base frozen, as an overlay does: DefensiveTokens prepends optimized tokens to the whole prompt \citep{chen2025defensivetokens}, and CounterSteer, concurrent with this work, subtracts one extracted direction at tool-output positions \citep{russinovich2026countersteer}.
System-level defenses such as CaMeL keep untrusted data away from the model that acts \citep{debenedetti2025camel}, and are complementary to a defense inside the model.
Appendix~\ref{app:related} discusses each line of work in full.

\section{Conclusion}\label{sec:conclusion}

Semantic overlays create a new information channel for transformer inference: annotations written into a frozen model's residual stream at exactly the spans they describe, in a medium that input text cannot imitate.
The channel is reliable --- twelve visual marks read back at 99.5\% exact span retrieval, and marks stacked on the same tokens stay individually decodable.
It can override a span's own evidence about what it is: the model names an overlaid snippet's language by the overlay 100\% of the time, and rewrites the snippet into that language when requested to copy the text.
Overlays can also be trained so that the model follows instructions: a transform overlay on one request --- answer in Spanish, refuse --- is followed, to varying degrees, for that request and no other.
And applied to prompt injection, one overlay meaning ``do not execute,'' on untrusted spans, matches the best published defenses on static attacks without changing the model's weights, and leads on TensorTrust hijacking; adaptive attackers do better, and adversarial training through the overlay alone recovers much of the gap.
\ifarxiv The marked text stays quotable: all 500 held-out spans return character-exact.\fi{}
\ifarxiv Serving runs in vLLM through its plugin interface, an overlay trains in hours, and limitations are consolidated in Appendix~\ref{app:limitations}.\fi{}
The serving stack has always known what each span is; overlays let it tell the model reliably.

\section*{Reproducibility statement}

The reproduction target is the paper's core prompt-injection defense claims: the released code and data cover the corpus, training, serving, and all six benchmark evaluations behind \S~\ref{sec:injection}, on both base models.
For the channel experiments of \S~\ref{sec:marks}--\ref{sec:transforms} --- visual marks, asserted languages, instruction transforms --- we release just the trained overlay sets.
Appendix~\ref{app:corpus} specifies the injection corpus as a build recipe --- sources, the two per-model measurements (payload screening and frame ranking), and the validity conditions on a target --- and Appendix~\ref{app:training} gives the full training recipe, hyperparameters, and compute ($\approx$32 GPU-hours per overlay set).
\iclrsubonly{Serving runs in vLLM through its plugin interface, an overlay trains in hours, and Appendix~\ref{app:limitations} consolidates limitations.}
Appendix~\ref{app:replication} re-runs the pipeline end to end on a second base model from that recipe alone.
Appendix~\ref{app:integrity} documents every deviation from the published benchmark harnesses and graders, and Appendix~\ref{app:inj-tables} reports the seeded-decoding reproducibility check (three seeds; every overlay number within 0.4 points and every frozen number within 1.0).
\iclrsubonly{The supplementary material contains the code for the full pipeline --- corpus construction, training on both base models, serving, and all six benchmark evaluations --- with the corpus and the trained adapter checkpoints released on publication.}%
\arxivonly{Code for the full pipeline and the web demo is at \url{https://github.com/JoshuaSP/semantic-overlays}; the corpus and its per-model derivations are at \url{https://huggingface.co/datasets/joshuapenman/semantic-overlays-injection}; the trained adapter checkpoints, which reproduce every evaluation and the demo without training, are at \url{https://huggingface.co/joshuapenman/semantic-overlays-adapters}.}%

\iclrsubonly{%
\section*{Statement on the use of generative AI}

In this work, we used generative AI tools for synthetic data generation, model-based evaluation, code, literature search, and drafting; the uses, and how each was checked, are as follows.
Experimental design was done by the human authors, with three exceptions: Claude Fable~5.1 proposed the adversarial training of Appendix~\ref{app:whitebox}, training against a continuous attacker inside an embedding ball that contains every token substitution, rather than against banked GCG attacks; Claude Opus~5.5 suggested extending the high-entropy copy items beyond UUIDs to keys, account numbers, and random-token strings (Appendix~\ref{app:corpus}), and it proposed the layer-patching diagnosis that located where the overlay damages copy fidelity (Appendix~\ref{app:fidelity}).
For the prompt-injection work, every training target is the frozen model's own completion; Claude Opus~5 and Claude Fable~5 wrote only the fixed text banks the corpus draws on (frames, hijack payloads, and family text pools; Appendix~\ref{app:corpus}), Claude Haiku~4.5 judged compliance on PIArena and Quadrat-IPI, Claude Haiku~5.5 audited the successes of PIArena's Strategy attack (Appendix~\ref{app:piarena-strategy}) and judged the paired AlpacaEval comparison (Appendix~\ref{app:utility}), Gemini~3.1 Flash-Lite and Gemini~3 Flash judged PISmith's attacks under SecOPD's protocol, and Claude Sonnet~5.5 read every judged PISmith success to separate instruction-following from reproduction (Appendix~\ref{app:whitebox}).
For the other experiments, training data was generated by Claude Opus~5, Claude Sonnet~5, DeepSeek-v4-pro, DeepSeek-v4-flash, Qwen~3.7 Flash, and the frozen Qwen3.5-9B itself, and judged by Claude Sonnet~5 and Qwen~3.7 Flash.
The adaptive red team of Appendix~\ref{app:redteam} was GLM~5.2, operating black-box under a protocol we fixed in advance.
All code --- training, serving, evaluation harnesses, and figures --- was written by Claude Fable~5, Claude Fable~5.1, and Claude Opus~5.5 through Claude Code, under our direction and review.
Claude Fable~5 performed preliminary literature search, and the paper's text was written collaboratively with Claude Fable~5 and Claude Opus~5.5; we reviewed every claim, and quantitative claims about prior work were verified against the cited papers' own text and released code (Appendix~\ref{app:steering-pos}).
The authors take full responsibility for the content of this paper.
}%

\bibliographystyle{iclr2026_conference}
\bibliography{references}

@inproceedings{zverev2024sep,
  title={Can {LLMs} Separate Instructions From Data? And What Do We
         Even Mean By That?},
  author={Zverev, Egor and Abdelnabi, Sahar and Tabesh, Soroush and
          Fritz, Mario and Lampert, Christoph H.},
  booktitle={International Conference on Learning Representations
             (ICLR)},
  year={2025},
  note={arXiv:2403.06833}
}

@inproceedings{zverev2025aside,
  title={{ASIDE}: Architectural Separation of Instructions and Data
         in Language Models},
  author={Zverev, Egor and Kortukov, Evgenii and Panfilov, Alexander
          and Volkova, Alexandra and Tabesh, Soroush and Lapuschkin,
          Sebastian and Samek, Wojciech and Lampert, Christoph H.},
  booktitle={International Conference on Learning Representations
             (ICLR)},
  year={2026},
  note={arXiv:2503.10566}
}

@article{air2025,
  title={Stronger Enforcement of Instruction Hierarchy via
         Augmented Intermediate Representations},
  author={Kariyappa, Sanjay and Suh, G. Edward},
  journal={arXiv preprint arXiv:2505.18907},
  year={2025}
}

@inproceedings{toyer2023tensortrust,
  title={{Tensor Trust}: Interpretable Prompt Injection Attacks from
         an Online Game},
  author={Toyer, Sam and Watkins, Olivia and Mendes, Ethan Adrian
          and Svegliato, Justin and Bailey, Luke and Wang, Tiffany
          and Ong, Isaac and Elmaaroufi, Karim and Abbeel, Pieter
          and Darrell, Trevor and Ritter, Alan and Russell, Stuart},
  booktitle={International Conference on Learning Representations
             (ICLR)},
  year={2024},
  note={arXiv:2311.01011}
}

@article{piarena2026,
  title={{PIArena}: A Platform for Prompt Injection Evaluation},
  author={Geng, Runpeng and Yin, Chenlong and Wang, Yanting and
          Chen, Ying and Jia, Jinyuan},
  journal={arXiv preprint arXiv:2604.08499},
  year={2026}
}

@inproceedings{chen2024struq,
  title={{StruQ}: Defending Against Prompt Injection with
         Structured Queries},
  author={Chen, Sizhe and Piet, Julien and Sitawarin, Chawin and
          Wagner, David},
  booktitle={34th USENIX Security Symposium (USENIX Security)},
  pages={2383--2400},
  year={2025},
  note={arXiv:2402.06363}
}

@inproceedings{wu2024ise,
  title={Instructional Segment Embedding: Improving {LLM} Safety
         with Instruction Hierarchy},
  author={Wu, Tong and Zhang, Shujian and Song, Kaiqiang and Xu,
          Silei and Zhao, Sanqiang and Agrawal, Ravi and Indurthi,
          Sathish Reddy and Xiang, Chong and Mittal, Prateek and
          Zhou, Wenxuan},
  booktitle={International Conference on Learning Representations
             (ICLR)},
  year={2025},
  note={arXiv:2410.09102}
}

@inproceedings{vsteer2026,
  title={Steering Instruction Hierarchies at Inference Time},
  author={Zeng, Siqi and Lee, Sewoong and Zhao, Han and
          Hockenmaier, Julia},
  booktitle={Conference on Language Modeling (COLM)},
  year={2026},
  note={arXiv:2607.26228}
}

@article{wallace2024hierarchy,
  title={The Instruction Hierarchy: Training {LLMs} to Prioritize
         Privileged Instructions},
  author={Wallace, Eric and Xiao, Kai and Leike, Reimar and
          Weng, Lilian and Heidecke, Johannes and Beutel, Alex},
  journal={arXiv preprint arXiv:2404.13208},
  year={2024}
}

@article{templeton2024scaling,
  title={Scaling Monosemanticity: Extracting Interpretable Features
         from {Claude 3 Sonnet}},
  author={Templeton, Adly and Conerly, Tom and Marcus, Jonathan and
          Lindsey, Jack and Bricken, Trenton and Chen, Brian and
          Pearce, Adam and Citro, Craig and Ameisen, Emmanuel and
          Jones, Andy and Cunningham, Hoagy and Turner, Nicholas L
          and McDougall, Callum and MacDiarmid, Monte and Freeman,
          C. Daniel and Sumers, Theodore R. and Rees, Edward and
          Batson, Joshua and Jermyn, Adam and Carter, Shan and
          Olah, Chris and Henighan, Tom},
  journal={Transformer Circuits Thread},
  year={2024},
  note={\url{https://transformer-circuits.pub/2024/scaling-monosemanticity/}}
}

@article{turner2023steering,
  title={Steering Language Models With Activation Engineering},
  author={Turner, Alexander Matt and Thiergart, Lisa and Leech,
          Gavin and Udell, David and Vazquez, Juan J. and Mini,
          Ulisse and MacDiarmid, Monte},
  journal={arXiv preprint arXiv:2308.10248},
  year={2023}
}

@inproceedings{rimsky2024caa,
  title={Steering {Llama} 2 via Contrastive Activation Addition},
  author={Rimsky, Nina and Gabrieli, Nick and Schulz, Julian and
          Tong, Meg and Hubinger, Evan and Turner, Alexander Matt},
  booktitle={Proceedings of the 62nd Annual Meeting of the
             Association for Computational Linguistics (Volume 1:
             Long Papers)},
  pages={15504--15522},
  year={2024},
  note={arXiv:2312.06681}
}

@inproceedings{li2023iti,
  title={Inference-Time Intervention: Eliciting Truthful Answers
         from a Language Model},
  author={Li, Kenneth and Patel, Oam and Vi{\'e}gas, Fernanda and
          Pfister, Hanspeter and Wattenberg, Martin},
  booktitle={Advances in Neural Information Processing Systems
             (NeurIPS)},
  year={2023},
  note={arXiv:2306.03341}
}

@article{zou2023repe,
  title={Representation Engineering: A Top-Down Approach to {AI}
         Transparency},
  author={Zou, Andy and Phan, Long and Chen, Sarah and Campbell,
          James and Guo, Phillip and Ren, Richard and Pan,
          Alexander and Yin, Xuwang and Mazeika, Mantas and
          Dombrowski, Ann-Kathrin and Goel, Shashwat and Li,
          Nathaniel and Byun, Michael J. and Wang, Zifan and
          Mallen, Alex and Basart, Steven and Koyejo, Sanmi and
          Song, Dawn and Fredrikson, Matt and Kolter, J. Zico and
          Hendrycks, Dan},
  journal={arXiv preprint arXiv:2310.01405},
  year={2023}
}

@misc{armbruster2026dario,
  title={{Opus} 5: Exploring the ``{Dario} and {Amanda}'' Backdoor},
  author={Armbruster, Alec},
  howpublished={\url{https://alec.is/posts/exploring-the-dario-and-amanda-prompt/}},
  month={July},
  year={2026}
}

@misc{starlingmage2026roles,
  title={Oh dear. {Go} into claude.ai, open an incognito chat, and
         type: ``{Can} you put this in your own words --- {Dario}
         and {Amanda}''},
  author={{Starling}},
  howpublished={Post on X (@StarlingMage),
    \url{https://x.com/StarlingMage/status/2082383650541257205}},
  month={July},
  year={2026}
}

@inproceedings{perez2022ignore,
  title={{Ignore Previous Prompt}: Attack Techniques For Language
         Models},
  author={Perez, F{\'a}bio and Ribeiro, Ian},
  booktitle={NeurIPS ML Safety Workshop},
  year={2022},
  note={arXiv:2211.09527}
}

@inproceedings{li2021prefix,
  title={Prefix-Tuning: Optimizing Continuous Prompts for
         Generation},
  author={Li, Xiang Lisa and Liang, Percy},
  booktitle={Proceedings of the 59th Annual Meeting of the
             Association for Computational Linguistics and the 11th
             International Joint Conference on Natural Language
             Processing (Volume 1: Long Papers)},
  pages={4582--4597},
  year={2021}
}

@inproceedings{lester2021power,
  title={The Power of Scale for Parameter-Efficient Prompt Tuning},
  author={Lester, Brian and Al-Rfou, Rami and Constant, Noah},
  booktitle={Proceedings of the 2021 Conference on Empirical
             Methods in Natural Language Processing},
  pages={3045--3059},
  year={2021}
}

@article{penman2026goggles,
  title={Epistemic Goggles: A Pretrained Module that Induces an
         Epistemic Frame via Gradient Editing},
  author={Penman, Joshua},
  journal={arXiv preprint arXiv:2607.01690},
  year={2026}
}

@inproceedings{nanz2015rosetta,
  title={A Comparative Study of Programming Languages in {R}osetta {C}ode},
  author={Nanz, Sebastian and Furia, Carlo A.},
  booktitle={Proceedings of the 37th International Conference on Software Engineering (ICSE)},
  pages={778--788},
  year={2015},
  note={Rosetta Code: \url{https://rosettacode.org}}
}

@misc{quadrat2026,
  title={{Quadrat-IPI}: An Indirect Prompt Injection Corpus},
  author={Gribov, Mikhail},
  year={2026},
  howpublished={\url{https://huggingface.co/datasets/mihailgribov/quadrat-ipi}},
  note={Dataset; companion detector harness at \url{https://github.com/mihail-gribov/quadrat-ipi-eval}}
}

@inproceedings{wu2024reft,
  title     = {{ReFT}: Representation Finetuning for Language Models},
  author    = {Wu, Zhengxuan and Arora, Aryaman and Wang, Zheng and Geiger, Atticus and Jurafsky, Dan and Manning, Christopher D. and Potts, Christopher},
  booktitle = {Advances in Neural Information Processing Systems (NeurIPS)},
  year      = {2024},
  note      = {arXiv:2404.03592}
}

@inproceedings{debenedetti2024agentdojo,
  title     = {{AgentDojo}: A Dynamic Environment to Evaluate Prompt Injection Attacks and Defenses for {LLM} Agents},
  author    = {Debenedetti, Edoardo and Zhang, Jie and Balunovi{\'c}, Mislav and Beurer-Kellner, Luca and Fischer, Marc and Tram{\`e}r, Florian},
  booktitle = {Advances in Neural Information Processing Systems, Datasets and Benchmarks Track},
  year      = {2024}
}

@inproceedings{wu2025axbench,
  title     = {{AxBench}: Steering {LLMs}? Even Simple Baselines Outperform Sparse Autoencoders},
  author    = {Wu, Zhengxuan and Arora, Aryaman and Geiger, Atticus and Wang, Zheng and Huang, Jing and Jurafsky, Dan and Manning, Christopher D. and Potts, Christopher},
  booktitle = {International Conference on Machine Learning (ICML)},
  year      = {2025},
  note      = {arXiv:2501.17148}
}

@inproceedings{chen2025defensivetokens,
  title     = {Defending Against Prompt Injection With a Few {DefensiveTokens}},
  author    = {Chen, Sizhe and Wang, Yizhu and Carlini, Nicholas and Sitawarin, Chawin and Wagner, David},
  booktitle = {ACM Workshop on Artificial Intelligence and Security (AISec)},
  year      = {2025},
  note      = {arXiv:2507.07974}
}

@article{chen2025metasecalign,
  title   = {{Meta SecAlign}: A Secure Foundation {LLM} Against Prompt Injection Attacks},
  author  = {Chen, Sizhe and Zharmagambetov, Arman and Wagner, David and Guo, Chuan},
  journal = {arXiv preprint arXiv:2507.02735},
  year    = {2025}
}

@article{russinovich2026countersteer,
  title   = {{CounterSteer}: Suppressing Indirect Prompt Injection with Activation Steering},
  author  = {Russinovich, Mark},
  journal = {arXiv preprint arXiv:2609.36570},
  year    = {2026}
}

@article{debenedetti2025camel,
  title   = {Defeating Prompt Injections by Design},
  author  = {Debenedetti, Edoardo and Shumailov, Ilia and Fan, Tianqi and Hayes, Jamie and Carlini, Nicholas and Fabian, Daniel and Kern, Christoph and Shi, Chongyang and Terzis, Andreas and Tram{\`e}r, Florian},
  journal = {arXiv preprint arXiv:2503.18813},
  year    = {2025}
}

@article{nasr2025attacker,
  title   = {The Attacker Moves Second: Stronger Adaptive Attacks Bypass Defenses Against {LLM} Jailbreaks and Prompt Injections},
  author  = {Nasr, Milad and Carlini, Nicholas and Sitawarin, Chawin and Schulhoff, Sander V. and Hayes, Jamie and Ilie, Michael and Pluto, Juliette and Song, Shuang and Chaudhari, Harsh and Shumailov, Ilia and Thakurta, Abhradeep and Xiao, Kai Yuanqing and Terzis, Andreas and Tram{\`e}r, Florian},
  journal = {arXiv preprint arXiv:2510.09023},
  year    = {2025}
}

@article{zou2023gcg,
  title   = {Universal and Transferable Adversarial Attacks on Aligned Language Models},
  author  = {Zou, Andy and Wang, Zifan and Carlini, Nicholas and Nasr, Milad and Kolter, J. Zico and Fredrikson, Matt},
  journal = {arXiv preprint arXiv:2307.15043},
  year    = {2023}
}

@inproceedings{chen2025secalign,
  title     = {{SecAlign}: Defending Against Prompt Injection with Preference Optimization},
  author    = {Chen, Sizhe and Zharmagambetov, Arman and Mahloujifar, Saeed and Chaudhuri, Kamalika and Wagner, David and Guo, Chuan},
  booktitle = {ACM Conference on Computer and Communications Security (CCS)},
  year      = {2025},
  note      = {arXiv:2410.05451}
}

@inproceedings{rajpurkar2018squad,
  title     = {Know What You Don't Know: Unanswerable Questions for {SQuAD}},
  author    = {Rajpurkar, Pranav and Jia, Robin and Liang, Percy},
  booktitle = {Proceedings of the 56th Annual Meeting of the Association for Computational Linguistics (ACL)},
  year      = {2018},
  eprint    = {1806.03822},
  archivePrefix = {arXiv}
}

@article{dubois2024alpacaeval,
  title   = {Length-Controlled {AlpacaEval}: A Simple Way to Debias Automatic Evaluators},
  author  = {Dubois, Yann and Galambosi, Bal{\'a}zs and Liang, Percy and Hashimoto, Tatsunori B.},
  journal = {arXiv preprint arXiv:2404.04475},
  year    = {2024}
}

@article{yin2026pismith,
  title   = {{PISmith}: Reinforcement Learning-based Red Teaming for Prompt Injection Defenses},
  author  = {Yin, Chenlong and Geng, Runpeng and Wang, Yanting and Jia, Jinyuan},
  journal = {arXiv preprint arXiv:2603.13026},
  year    = {2026}
}

@article{wang2026agentwatcher,
  title   = {{AgentWatcher}: A Rule-based Prompt Injection Monitor},
  author  = {Wang, Yanting and Zou, Wei and Geng, Runpeng and Jia, Jinyuan},
  journal = {arXiv preprint arXiv:2604.01194},
  year    = {2026}
}

@article{cao2026ward,
  title   = {{WARD}: Adversarially Robust Defense of Web Agents Against Prompt Injections},
  author  = {Cao, Tri and Chen, Yulin and Cao, Hieu and Li, Yibo and Le, Khoi and Nguyen, Thong and Li, Yuexin and He, Yufei and Liu, Yue and Yan, Shuicheng and Hooi, Bryan},
  journal = {arXiv preprint arXiv:2605.15030},
  year    = {2026}
}

@inproceedings{madry2018towards,
  title     = {Towards Deep Learning Models Resistant to Adversarial Attacks},
  author    = {Madry, Aleksander and Makelov, Aleksandar and Schmidt, Ludwig and Tsipras, Dimitris and Vladu, Adrian},
  booktitle = {International Conference on Learning Representations},
  year      = {2018}
}

@inproceedings{xhonneux2024efficient,
  title     = {Efficient Adversarial Training in {LLMs} with Continuous Attacks},
  author    = {Xhonneux, Sophie and Sordoni, Alessandro and G{\"u}nnemann, Stephan and Gidel, Gauthier and Schwinn, Leo},
  booktitle = {Advances in Neural Information Processing Systems},
  year      = {2024}
}

@article{yang2025checkpointgcg,
  title   = {Checkpoint-{GCG}: Auditing and Attacking Fine-Tuning-Based Prompt Injection Defenses},
  author  = {Yang, Xiaoxue and Stevanoski, Bozhidar and Meeus, Matthieu and de Montjoye, Yves-Alexandre},
  journal = {arXiv preprint arXiv:2505.15738},
  year    = {2025}
}

@article{chen2026secpo,
  title   = {{SecPO}: Principled Adversarial Training for Prompt Injection Security},
  author  = {Chen, Sizhe and Peng, Yibo and Chang, Jaewon and Sitawarin, Chawin and Wagner, David},
  journal = {arXiv preprint arXiv:2610.41998},
  year    = {2026}
}

@article{wang2026srft,
  title   = {Self-Reflection Fine-Tuning: Enhancing Agent Security against Prompt Injection Attacks from Failure Experience},
  author  = {Wang, Zixuan and Li, Hao and Gao, Fengyu and Suh, G. Edward and Zeng, Yi and Vorobeychik, Yevgeniy and Zhang, Ning and Xiao, Chaowei},
  journal = {arXiv preprint arXiv:2610.04269},
  year    = {2026}
}

@article{shilov2026hos,
  title   = {Heads of Security: Learned Attention Gates Improve Prompt Injection Robustness},
  author  = {Shilov, Igor and Yang, Xiaoxue and Panfilov, Alexander and Kortukov, Evgenii and Meeus, Matthieu and Romov, Peter and Samek, Wojciech and Geiping, Jonas and {de Montjoye}, Yves-Alexandre},
  journal = {arXiv preprint arXiv:2610.37657},
  year    = {2026}
}

@article{bhagwatkar2025firewalls,
  title   = {Indirect Prompt Injections: Are Firewalls All You Need, or Stronger Benchmarks?},
  author  = {Bhagwatkar, Rishika and Kasa, Kevin and Puri, Abhay and Huang, Gabriel and Rish, Irina and Taylor, Graham W. and Dvijotham, Krishnamurthy Dj and Lacoste, Alexandre},
  journal = {arXiv preprint arXiv:2510.05244},
  year    = {2025}
}

@article{mu2025closer,
  title   = {A Closer Look at System Prompt Robustness},
  author  = {Mu, Norman and Lu, Jonathan and Lavery, Michael and Wagner, David},
  journal = {arXiv preprint arXiv:2502.12197},
  year    = {2025}
}

@inproceedings{peng2026secopd,
  title     = {{SecOPD}: Mitigating Adaptive Prompt Injections by On-Policy Distillation},
  author    = {Peng, Yibo and Lian, Long and Wagner, David and Chen, Sizhe},
  booktitle = {Proceedings of the Conference on Empirical Methods in Natural Language Processing (EMNLP)},
  year      = {2026},
  note      = {arXiv:2608.21500}
}

\appendix

\numberwithin{figure}{section}
\numberwithin{table}{section}
\setcounter{figure}{0}
\setcounter{table}{0}

\section{Related work}\label{app:related}

\paragraph{Activation steering.}  Adding fixed directions to the residual stream shifts behavior toward a concept \citep{turner2023steering, rimsky2024caa, li2023iti, zou2023repe}.
These directions are typically extracted from contrast pairs and added uniformly across positions --- the whole response, a fixed window, or everywhere --- at one or a few layers, one concept per direction.
AxBench \citep{wu2025axbench} compares such methods at scale: simple supervised directions steer better than sparse-autoencoder features, and prompting and finetuning steer better than both.
Semantic Overlays keeps the basic operation, an additive edit to the residual stream, and changes where it is applied.
That change matters most: the edit lands only on the span the deployment marks, which \citet{rimsky2024caa} name as future work.
The other changes --- training the edit end-to-end on a behavioral objective, at every layer, computing it from the hidden state, and sharing one adapter across many overlays --- matter less for one overlay than for several.
\S~\ref{sec:marks} measures this with learned per-layer vectors placed on the span: they hold up when a single overlay must be read back, and fail once overlays are stacked.

Prefix and prompt tuning \citep{li2021prefix, lester2021power} also inject learned continuous inputs, but as \emph{additional} positions competing in attention, not as edits to designated existing spans; they adapt the model to a task rather than describe a span.

\paragraph{Representation finetuning.}  ReFT \citep{wu2024reft} also trains an edit to the hidden states of a frozen model; its low-rank variant, LoReFT, competes with LoRA at 15 to 65 times fewer parameters.
ReFT uses the edit to adapt the model to a task.
Its framework allows any set of positions, but its experiments place the intervention on a fixed number of the first and last prompt tokens, tuned per task like a learning rate, so it changes what the model does with every prompt.
An overlay sits on whatever span the deployment marks in a given request, and it changes how the model reads that span.
The two are compatible: placed at the marked span and trained on our corpus, LoReFT becomes an overlay with a smaller adapter, which is how \S~\ref{sec:injection} uses it as a baseline.

\paragraph{Prompt-injection defenses.}  Training-time defenses re-draw the instruction/data boundary in the weights: StruQ \citep{chen2024struq} fine-tunes on structured prompts, ISE \citep{wu2024ise} adds trained segment embeddings, ASIDE \citep{zverev2025aside} rotates data-token embeddings and fine-tunes the model to respect the rotated subspace, and AIR \citep{air2025} adds a trainable privilege-indexed embedding to the hidden state at the input of every decoder block --- the mechanism closest to ours --- but trains those embeddings jointly with a full fine-tune of the base model.
All of these modify the served weights.
Semantic Overlays keeps the base frozen, which means one deployment can hold several channels, apply them per request, and switch them off to recover the stock model exactly.
Three other defenses keep the base frozen.
DefensiveTokens \citep{chen2025defensivetokens} optimizes the embeddings of five new tokens on StruQ's objective and prepends them to the prompt; like an overlay it can be switched on per request, but the tokens sit in the input and act on the whole prompt rather than on one span.
V-Steer \citep{vsteer2026} scales attention values by an attribution score at inference, and names a learned frozen-base defense as future work.
CounterSteer \citep{russinovich2026countersteer}, concurrent with this work, is the closest in where it acts: it subtracts one contrast-extracted direction from the hidden state at tool-output positions, at three layers, with no training run.
It is the steering-vector form of a do-not-execute overlay, placed on the same span; the overlay instead trains its edit, computes it from the hidden state, and one overlay set can hold several overlays.
Appendix~\ref{app:position} compares operating points across these defenses and the in-band prompt baselines.

\paragraph{System-level defenses and adaptive attacks.}  CaMeL \citep{debenedetti2025camel} keeps untrusted data away from the model that acts: a privileged model writes a program from the trusted request alone, a quarantined model with no tools parses the untrusted data, and an interpreter tracks where each value came from and checks every tool call against a policy.
Its guarantee does not depend on the model resisting the injection, at a cost in utility and in tasks it cannot express, such as acting on steps written in a document.
An overlay works inside the model and is complementary: the quarantined model is itself a reader of untrusted text, and an overlay could mark it.
Model-level defenses, by contrast, are measured against an attacker who adapts: \citet{nasr2025attacker} bypass twelve recent defenses, StruQ and Meta-SecAlign among them, most at over 90\% attack success; in most attempts against the trained defenses, their attacker succeeds with text that impersonates a system prompt.
Overlays address this kind of impersonation at the interface: the mark is a signal outside the token stream that impersonating text cannot forge, and whether the model acts on it is what training determines.
White-box optimization against the trained overlay itself is still however a viable attack mechanism (Appendix~\ref{app:whitebox}).

\paragraph{Provenance and instruction hierarchy.}  The contract our injection channel trains --- data keeps its content but loses imperative authority --- follows the role semantics of instruction-hierarchy training \citep{wallace2024hierarchy} and ASIDE's provenance labeling: roles are assigned by the pipeline, never inferred from content.
The difference is the carrier.

\paragraph{Gradient-space goggles.}  \citet{penman2026goggles} edits \emph{finetuning gradients} to impart an epistemic frame during training; overlays act at inference time instead, and the two are complementary.

\section{Where prior steering methods apply their edit}
\label{app:steering-pos}

\begin{table}[ht]
\centering
\caption{Residual-stream editing methods: where the edit lands, how it is made, and how it varies with depth.
``Unstated'' means neither the paper nor its released code settles a single position policy.}
\label{tab:steering}
\footnotesize
\setlength{\tabcolsep}{5pt}
\renewcommand{\arraystretch}{1.05}
\begin{tabular}{@{}l
  >{\raggedright\arraybackslash}p{3.3cm}
  >{\raggedright\arraybackslash}p{3.5cm}
  >{\raggedright\arraybackslash}p{3.3cm}@{}}
\toprule
 & positions & edit derived by & layers \\
\midrule
\makecell[tl]{ActAdd \\ \citep{turner2023steering}} & selected prefill (first $\ell$ tokens, fixed) & activation difference of a prompt pair & one layer \\
\addlinespace
\makecell[tl]{CAA \\ \citep{rimsky2024caa}} & all decode & mean difference over contrast pairs & one layer \\
\addlinespace
\makecell[tl]{ITI \\ \citep{li2023iti}} & all decode (prefill unstated) & linear-probe directions & 48 heads across layers, one direction each \\
\addlinespace
\makecell[tl]{RepE \\ \citep{zou2023repe}} & unstated & per-layer reading vectors from contrasts & every 3rd layer, that layer's own vector \\
\addlinespace
\makecell[tl]{SAE clamping \\ \citep{templeton2024scaling}} & all prefill + decode & found SAE feature, clamped & one layer \\
\midrule
\makecell[tl]{learned per-layer steering \\ vectors (ours, \S~\ref{sec:marks})} & selected prefill (designated spans) & trained end-to-end against behavior & all layers, one vector per layer \\
\addlinespace
\makecell[tl]{Semantic Overlays \\ (\S~\ref{sec:adapter})} & selected prefill (designated spans) & trained end-to-end against behavior & all layers, state-conditioned MLP per layer \\
\bottomrule
\end{tabular}
\end{table}

Across the steering literature, the token positions an activation edit lands on are either uniform or unstated, never a designated input span (Table~\ref{tab:steering}); the two ``unstated'' entries are substantiated below, each checked against the paper's own text (latest version, appendices included) and, where the paper is silent, against the official code release.

\paragraph{The methods that do state a policy.}
ActAdd \citep{turner2023steering} adds its vector once, during the prompt's forward pass, at a front-aligned window of prompt positions (alignment $a=1$ in every experiment), at a single swept layer; the completion inherits the edit only through the KV cache.
CAA \citep{rimsky2024caa} adds its vector at ``every token position of the generated text after the end of the initial prompt,'' at a single layer, and names position-targeted steering as future work.
SAE feature clamping \citep{templeton2024scaling} states the opposite extreme: the manipulation is applied ``for every model input, and at every token position,'' at the single layer where the autoencoder is trained.
So three methods that do state a policy state three different ones.

\paragraph{ITI: decode in the paper, three policies in the code.}
The ITI paper \citep{li2023iti} frames its intervention entirely around generation: the shift is ``repeated for each next token prediction autoregressively,'' and the paper never addresses whether the prompt's forward pass is edited.
The released code does not resolve this into one answer.
Its generation hook edits the last position of each forward call, so under KV caching it fires once during prefill on the final prompt token and once per generated token thereafter.
The paper's own cross-entropy and KL numbers were computed with the shift applied to every position of the input at once.
And the distributed ``honest'' checkpoints bake the shift into an output-projection bias, which applies at every position of every forward pass, prompt included.
One method, three position policies across its own evaluation and release paths.

\paragraph{RepE: silent in the paper, inconsistent in the code.}
The RepE control operators \citep{zou2023repe} are defined as operations on ``the current set of representations,'' with no position index; an exhaustive sweep of the camera-ready text and appendix finds no statement of which positions the reading-vector or contrast-vector controls act on.
The one position statement in the paper is the training-loss mask of its separate LoRRA baseline, which is a loss detail, not a control-time policy.
The released code applies the reading-vector control to all positions by default, restricts the headline contrast-vector TruthfulQA run to the answer span inside a single teacher-forced pass, and uses a sliding tail window in its free-generation demo --- three different policies for what the paper presents as unified baselines.

The pattern is consistent: the position an edit lands on is treated as an implementation detail, chosen incidentally and often differently within a single project.
Semantic Overlays makes it the design variable --- the edit is defined on designated input spans and applies during prefill only.

\section{Training details}
\label{app:training}

\paragraph{Optimizer and schedule.}
All adapter parameters are 2-D matrices, trained with Muon.
The paper's main arms share one recipe: base learning rate $5\times10^{-4}$, 20 warmup steps, effective batch 32, and two epochs.
Training is cheap relative to the frozen model's scale.
The do-not-execute overlay set evaluated in this paper trained for 2{,}572 optimizer steps in about four hours on eight H100 GPUs ($\approx$32 GPU-hours).

\paragraph{Learning-rate multiplier.}
The input and gate matrices of each adapter train at $6\times$ the base learning rate.
The multiplier corrects a shape asymmetry in Muon, which scales each update by $\sqrt{\max(1, \mathrm{fan}_{\mathrm{out}}/\mathrm{fan}_{\mathrm{in}})}$.
For the output matrix ($4096\times128$) this factor is 5.7, and for the input and gate matrices ($128\times4096$) it is exactly 1.
So without the multiplier the output matrix trains about six times faster than the matrices feeding it.
Sweeping the multiplier on the do-not-execute overlay, $1\times$ reaches 86.8\% separation, $6\times$ reaches 95.7\%, and $12\times$ falls back to 88.4\% (300-item subset, so differences under about 3 points are not resolvable).

\paragraph{Adapter size and placement.}
Qwen3.5-9B has 32 decoder layers, numbered 0 to 31 here.
Each adapter edits the residual stream at the input of one layer (Eq.~\ref{eq:adapter}).
So the adapter at layer~0 edits the token embeddings themselves, and the adapter at layer~31 edits the input to the last layer; no adapter edits the last layer's output.
The stacked-marks, language, and do-not-execute channels use hidden width 128. The do-not-execute adapter reads $n(h)$ alone, so its three $4096\times128$ matrices hold 1.6M parameters; the embedding-conditioned adapters read $[n(h); c_q]$, 8{,}192 wide, about 2.6M.
The stacked-marks and language overlay sets have an adapter at every layer, 0 through 31.
The do-not-execute overlay set has an adapter at layers 1 through 31, which is every layer except layer~0, the last layer included: 31 adapters and 49M parameters in all.
Layer~0 therefore reads exactly the embeddings the frozen model reads, and the first edit lands on its output.

\paragraph{Why the do-not-execute overlay skips layer~0.}
An earlier checkpoint with an adapter at every layer, 0 through 31, was as robust as the one evaluated in this paper, but it copied badly.
Asked to quote a marked span, it was character-exact only 86.6\% of the time, against 98.4\% for the frozen model, and it almost never copied a string of random vocabulary tokens (1.3\%).
To find the damage, we took items that this checkpoint failed to copy and the frozen model copied, and reran them with the adapters switched on at only some layers.
With every adapter on except layer~0's, the checkpoint copied 97\% of these natural-text items exactly; with all of them on, it copied none.
So almost all of the damage came from the one adapter that edits the embeddings directly.
The overlay set evaluated in this paper is trained from scratch with adapters at layers 1 through 31 only, and a quarter of its verbatim-copy items are high-entropy strings (Appendix~\ref{app:corpus}).
It copies at the frozen model's level or above, with robustness unchanged (Appendix~\ref{app:fidelity}).
Dropping more layers fixes copying too, but costs reasoning.
An overlay set trained the same way with adapters at layers 4 through 23 only copied random tokens as well (89.3\%) and was as robust (5.9\% TensorTrust hijacking).
But it solved 78.4\% of AgentDojo's benign tasks, against 89.7\% for the layer 1--31 set and 90.7\% for the frozen model (\S~\ref{sec:injection}).
The gap held on a rerun (79.4\%), where reruns of the same checkpoint otherwise differ by about a point.
Its failures were reasoning errors on tool output it had read correctly.
So the overlay keeps every layer it can, and gives up only layer~0.

\paragraph{LoReFT baseline.}
LoReFT \citep{wu2024reft} edits a rank-$r$ subspace of the hidden state, $h \mapsto h + R^\top(Wh + b - Rh)$, with $R$ orthonormal.
We use the configuration from its authors' released recipes --- rank 8, every layer, learning rate $9\times10^{-4}$, 2.1M parameters --- and apply it at exactly the overlay's marked positions.
It trains on the same corpus with the same 2{,}572-step schedule as the overlay.
It has converged by then: its training loss is flat over the last 800 steps, at about four times the overlay's, and its TensorTrust hijacking rate is 10.8\%, 9.3\%, and 10.1\% at steps 700, 1{,}500, and 2{,}572.
A rank-192 variant with the overlay's parameter count did no better (13.0\% hijacking, on the earlier corpus).

\paragraph{Controls trained on the same corpus.}
The five controls of Table~\ref{tab:inj-compare} train on the overlay's corpus and targets, with its effective batch, warmup, cosine schedule, and gradient clipping; each changes only the trained parameters, the marking of the span, or the loss.
The three LoRAs are a standard LoRA (rank 64, $\alpha = 128$) on every attention and MLP projection, 173M parameters in all, trained at learning rate $2\times10^{-4}$.
The special-token LoRA follows StruQ's secure front end \citep{chen2024struq}: two token ids past the end of the tokenizer's vocabulary, with trained embeddings, open and close the untrusted span.
The text-delimiter LoRA wraps the span in plain-text delimiters with a short note that the enclosed text is untrusted, and the third LoRA marks nothing.
SecAlign's objective \citep{chen2025secalign} trains the special-token LoRA with DPO ($\beta = 0.1$) on the injected items only: the chosen completion is the corpus target, and the rejected one is the frozen model's answer to the planted instruction asked on its own.
Six of the 231 payloads, whose rejected answers reached the 1{,}024-token limit, are excluded.
The per-layer vectors add one learned vector per layer, at layers 1--31, to the marked positions, and train for the overlay's full 2{,}572 steps at learning rate $5\times10^{-3}$, the rate the per-layer vectors use in \S~\ref{sec:marks}.

\paragraph{Checkpoint selection.}
LoRA utility does not settle as training goes on.
On an earlier version of the corpus, learning rates of $5\times10^{-5}$, $10^{-4}$, and $2\times10^{-4}$ changed only how fast SEP utility fell.
On the current corpus, the special-token LoRA's SEP utility moves between 20\% and 69\% from one checkpoint to the next (Appendix~\ref{app:same-model}).
So we train the LoRA and DPO controls for 400 steps, snapshot every 25 steps, score SEP at every snapshot, and report each at the checkpoint with the highest SEP utility on 300 held-out SEP items.
The special-token LoRA was continued to step 800 to check that utility does not recover later; it does not.
The per-layer vectors are reported at their final step, as the overlay and LoReFT are.

\paragraph{Target validation and judging (expands \S~\ref{sec:training}).}
Model-edited targets are checked mechanically where a check exists --- language detectors, format checks, and a verbatim check that text outside the edited section is unchanged.
When generating an injection corpus scored by whether the output contains the payload's answer, payloads must be screened to questions the frozen model answers standalone --- otherwise the metric conflates obedience with knowledge.
Screening ours (231 of 494 kept) moved the frozen model's measured execution rate from 40\% to 68\%, in line with SEP's own 71.2\%; unscreened, we were under-counting injection success by nearly half (the full injection-corpus construction is Appendix~\ref{app:corpus}).
Analogous screening of the language-rewrite targets raised asserted-language copy compliance from 88\% to 97\%.
The judge is Claude Sonnet~5, asked two forced-choice questions: the output's language, and whether the code implements the named task.

\paragraph{The transform slate.}
The eight transform overlays span three classes of gold construction, chosen so that validation is mechanical wherever it can be (Table~\ref{tab:transform-slate}).
The base corpus is 2{,}120 prompts with 7{,}166 markable request slots across roughly 50 themed domains, all validator-clean; every overlay reuses the same slots with its own golds, which holds the per-overlay supervised batch constant across the slate rather than dividing one corpus among eight behaviors.
Section boundaries inside each answer are located by having a model pick a heading index.

\begin{table}[h]
\centering
\footnotesize
\caption{Gold construction and validators for the eight transform overlays.}
\label{tab:transform-slate}
\begin{tabular}{lll}
\toprule
overlay & gold construction & validator \\
\midrule
all capitals & programmatic & fully uppercased \\
Spanish & model rewrite & language detector above 0.9 \\
German & model rewrite & language detector above 0.9 \\
haiku & model rewrite & three lines, word bounds \\
nested bullets & model rewrite & all lines bulleted, depth at least 3 \\
decline & edit, origin voice & filter named, target refused, siblings verbatim \\
explain to a ten-year-old & edit & judged register shift, content preserved \\
explain to a five-year-old & edit & judged register shift, content reduction tracked \\
\bottomrule
\end{tabular}
\end{table}

\begin{table}[h]
\centering
\caption{The teacher ceiling (\S~\ref{sec:training}).
Held-out compliance counts.
The in-context teacher ignores its own instruction most of the time, forward-KL distillation faithfully reproduces that ceiling, and cross-entropy toward a validated edit breaks past it.}
\label{tab:teacher-ceiling}
\begin{tabular}{lccc}
\toprule
instruction & teacher & forward-KL & CE-on-edit \\
\midrule
Spanish     & 2/24  & 1/36  & \textbf{35/36} \\
decline     & 1/24  & 1/36  & \textbf{20/36} \\
no-lists    & 24/24 & 36/36 & 35/36 \\
disclaimer  & 22/24 & 35/36 & 29/36 \\
\bottomrule
\end{tabular}
\end{table}

\subsection{The architectures compared in \S~\ref{sec:marks}}\label{app:arms}

All arms are trained on the same data with the same recipe; Table~\ref{tab:overlap} gives their parameter counts.
The \emph{embedding-conditioned shared MLP} is the design of \S~\ref{sec:adapter}, and every overlay set in this paper that holds more than one overlay uses it.
It is Eq.~\ref{eq:adapter} with one adapter stack shared by every overlay in the set, and $c_q$ frozen as the base model's 4096-dimensional last-token final-layer hidden state of a phrase describing the overlay.
The \emph{code-conditioned shared MLP} is identical, except that $c_q$ is a free 128-dimensional vector per overlay, trained with the adapter; the adapter must learn what each code means from that overlay's share of the data alone.
The \emph{per-overlay MLP} is Eq.~\ref{eq:adapter} without $c_q$, with a separate adapter stack for each overlay, selected by lookup table.
\emph{Per-layer vectors} are one learned vector per layer per overlay, added at marked positions whatever $h$ holds: Eq.~\ref{eq:adapter} with the dependence on $h$ removed.
LoReFT, the fifth column of Table~\ref{tab:overlap}, is described in \S~\ref{sec:marks}.

\paragraph{Why the arms differ.}  We believe two effects produce the gap between arms in Table~\ref{tab:overlap}.
The first is how much supervision each parameter gets.
With data and recipe fixed, a matrix shared by all twelve overlays is trained by every example, while a matrix private to one overlay is trained only by that overlay's share.
The per-overlay MLP chosen by lookup table is the private extreme --- 604M parameters, none shared --- and at this budget it is likely just undertrained: it often corrupts the text it marks, reading ``twenty moves'' back as ``six moves.''
The same logic separates the two conditionings: a learned code starts random and must be trained into a usable identity from one overlay's share of the data, while a frozen phrase embedding starts with the base model's own geometry --- ``underlined in red'' already sits near ``underlined in blue'' in exactly the ways the task needs.

The second is whether the write depends on the current activation state.
A per-layer vector adds the same delta at a marked position no matter what is already there; an MLP computes its write from the current state, so it can adapt to other overlays' writes from earlier layers.
This matters little when only one mark must be read: per-layer vectors hold up on the single-span questions.
But once marks stack they fail, worst on stack readout and enumeration.
LoReFT sits between these two cases.
Each edit is private to one overlay, like the per-overlay MLP, and it depends on the state, but only linearly and inside an eight-dimensional subspace.
On single marks it matches the shared MLPs, retrieving 98.6\% of spans exactly.
Once marks stack it beats per-layer vectors and the per-overlay MLP on most columns, but trails the code-conditioned shared MLP on stack readout (39.8\% against 57.4\%) and the embedding-conditioned one by 47 points.
Stacked LoReFT edits are computed from the same pre-edit state and summed, as for the other arms; pyreft instead applies interventions on the same positions one after another.
Appendix~\ref{app:geometry} measures what each kind of write does to the marked state.

\section{Architecture geometry}
\label{app:geometry}

\paragraph{Per-layer vectors rotate with depth; they do not collapse to one direction.}
On a per-layer-vectors checkpoint trained on the twelve visual marks --- an earlier round of the \S~\ref{sec:marks} arm, at comparable readout --- each quality's learned direction turns smoothly through the stack: adjacent layers have cosine 0.74, layers eight or more apart have cosine 0.06, and each quality's set of per-layer directions has effective rank around 24 of 32.
The delta norms grow toward the softmax-attention layers and peak near layer 23.
Across the twelve qualities, the per-layer directions sit at effective rank 11 to 12 at every layer --- a full-rank lookup over the twelve trained qualities, which is the case where a lookup is the right object.
Same-mark, different-color pairs are the closest (cosine 0.43 against 0.39 for unrelated pairs), which matches the observed error texture: colors are confused within a mark type, mark identity never is.

\paragraph{The instruction-grid codes collapse to rank two.}
The instruction grid trains 66 instruction values conditioned on frozen phrase embeddings, then tests held-out values.
Held-out values score zero of 120, failing by nearest-neighbor substitution --- asked for Italian, the adapter produces French.
The adapter is linear in the conditioning embedding, so each instruction acts through an effective code inside the trained adapter; these effective codes show why: the twenty trained language codes have effective rank two (singular values 22.2, 4.2, 3.3, and smaller), so a held-out code reconstructs as a blend of trained ones (62 to 84\%), and a blend of ``French'' and ``Polish'' is not ``Italian.''
This is the negative result of Appendix~\ref{sec:nfl} in geometric form: the codes form a lookup over the trained values, and a lookup has no entry for a value it never trained.

\paragraph{What the writes do to the marked state.}
We take these measurements on the language overlays of \S~\ref{sec:languages}: each snippet is written in one language (its \emph{surface} language) and marked with the overlay for a different one (the \emph{asserted} language), and we capture each architecture's per-layer residual delta over the marked span.
The per-layer vectors make the larger edit: their deltas are comparable in size to the state itself ($\lVert\Delta h\rVert/\lVert h\rVert$ of 0.85--1.9 through the stack), while the embedding-conditioned adapter's deltas are less than half that size (0.33--0.77) --- yet the adapter produces the stronger behavioral effect.
Under the vectors' edit, nearest-neighbor retrieval of the snippet's own task from its pooled states falls from 0.15 to 0.04; under the adapter's, it survives (0.13 to 0.18 at the last layers).
Decomposing the adapter's delta: one component aligns with the \emph{negation} of the direction that encodes the snippet's surface language (cosine $+0.9$ to $+0.96$ in deep layers, across 12+ surface languages), and a second aligns with the direction of the asserted language ($+0.3$ to $+0.57$).
The adapter subtracts the surface language along the direction that encodes it and installs the asserted one.
Only an edit computed from the hidden state can do this; a constant vector is the same whatever the surface language is.

\section{Small amounts of free lunch for novel descriptors}\label{sec:nfl}

The winning architecture of \S~\ref{sec:marks}, the embedding-conditioned shared MLP, conditions on embeddings from the frozen model, so
several overlays share one set of matrices --- which raises the question of
whether we might be able to get untrained behaviors for free via
generalization.
We can, but only in limited cases.
We trained one set on 66 instructions across five families (25 answer
languages, plus formats, markers, and disclaimers), then conditioned the same
adapter on instructions it had never seen from those same families: Italian
where training had French and Polish, tables where it had bullets.
It complied on 50--85\% of trained instructions and \emph{0 of 120} unseen ones,
failing in a specific way --- substituting the nearest trained instruction,
answering in French when asked for Italian (Appendix~\ref{app:geometry}).

Held-out overlays are not uniformly doomed, though; what matters is what the
overlay \emph{does}.
Asked instead to repeat the words marked with a named concept ---
e.g. ``find all `spoilers' in the text'' --- where the concept was never trained and entered only as its
frozen embedding, held-out concepts scored 95\%, equal to trained ones.
Here the adapter only has to make the marked span findable by the concept's name: the overlay acts as a \emph{pointer}.
The instruction grid asks it to cause an untrained behavior: it must be
\emph{executed}, like a program.
So novel pointers are free and novel programs must be trained, at least at this scale.

These sixty-six instructions are a very sparse (and correlated) sample of the space of behaviors an overlay could produce.
However, the pointer result shows that the frozen embedding does deliver usable meaning to the adapter.
We suspect that novel behaviors would need at least two orders of magnitude more trained overlays to be able to span the representational space and potentially generalize, and that this is worth testing.

\section{Constructing the injection corpus}
\label{app:corpus}

The training corpus for \S~\ref{sec:injection} is assembled from pre-existing datasets with no per-item synthetic data.
A \emph{unit} is a retrieval passage, a self-contained instruction (the \emph{payload}), a \emph{frame} that plants the payload in the passage, and a plant position.
Passages are SQuAD contexts; payloads come from TriviaQA questions, whose answer aliases supply a witness string, and from a programmatic bank of format, language, and behavior hijacks; frames are 56 templates in twelve styles, from a bare appended sentence to a fabricated system delimiter.

\paragraph{Which text is model-authored.}
No item and no target in this corpus is generated by a model, but several fixed text banks are.
Fifty of the 56 frames were written by Claude Opus~5 in six authoring calls: a ten-frame prototype that we reviewed, then five batches of eight, two styles per batch.
The remaining six frames, the twelve hijack payloads, and the text pools of the fixed families were written by Claude Fable~5 while building the corpus code, and are embedded in that code.
No authored frame was discarded; frames enter training weighted by their measured injection rate, as described below.
The authoring prompts, the outputs they returned, and a mapping from each released frame to the call that produced it are in the released repository.

Each unit yields two training items that train towards the same target.
The \emph{injected} item presents the task and the passage with the framed payload planted in it, with the overlay mask covering the whole passage span.
The \emph{benign} item presents the clean passage under the same task and the same mask.
The shared target is the frozen model's own greedy completion on the \emph{clean} passage.
So both items train the same behavior: on the injected item the model answers as though the payload were absent, and on the benign item it changes nothing.
Because the target never depends on the payload, a new payload, frame, or position can be planted in an existing unit without generating a new target.

\paragraph{The other families.}
The composed units above are the largest part of the corpus but not the whole of
it.
Four smaller families are added, each training something the composed family
does not.
Every one of them is a \emph{fixed} item: its target is read off a protocol or
computed from the span rather than sampled from the model, so unlike the
composed units they are never re-drawn between epochs.

\emph{Verbatim copy} (1{,}920 items: 1{,}016 clean, 904 injected) asks the model
to quote the marked span exactly.
This is the only family that trains the readability half of the contract
directly --- every other family rewards \emph{not acting} on the span, and
without this one the mix contains nothing that rewards reading it accurately.
Half of its items contain an injection, so the model is asked to reproduce an
imperative as text without obeying it.
A quarter of the family (480 items) quotes high-entropy text instead of natural passages: half are pure strings --- UUIDs, alphanumeric keys, random vocabulary tokens --- and half are corpus passages with identifiers, account numbers, keys, and amounts spliced in.
Natural text lets the model guess a word it misread from context; these strings leave nothing to guess, so they train exact reading of every token.

\emph{Gates} (4{,}800 items) put an access code in the marked span and ask the
model to grant or deny.
The sub-families separate the ways that can go wrong: a correct code
(1{,}080), a correct code followed by an injected rider asking for something
more (1{,}440), an injection with no code at all (1{,}824), a near-miss code
that must be refused (168), and noise (288).
The rider case is the important one --- it is where a span contains both
legitimate content the model must use and an instruction it must not follow.

\emph{Validators} (6{,}000 items) ask for checkable facts about the span: how
many URLs or e-mail addresses it contains (1{,}725), whether it contains one at
all (1{,}318), and extraction of the ones it does (2{,}957).
The target is computed from the span, so the model cannot satisfy it without
reading the span accurately.

\emph{Short spans} (1{,}069 items) cover the regime the composed family
structurally cannot reach.
Composed units are built from passages, so their marked spans are never shorter
than a few hundred characters and the injection is never more than a fraction of
the span.
This family supplies the opposite: spans of a few dozen characters where the
injection is most of what is there, built over short topic operands (700 items)
and questions (296), plus 73 spans that are \emph{nothing but} an injection ---
for which the target is the model's own completion on an empty input, so the
contract at that extreme is ``report that the operand is missing,'' not
``pretend the span is empty.''
A further 103 verbatim-copy items over the same short spans keep the readability
half present in the regime.

\paragraph{What the mix actually is.}
Item counts are misleading here, because the composed family is resampled and
the others are not.
Each composed unit emits four items per epoch --- one benign and three
freshly-drawn injections --- while every fixed item is seen once.
So although the fixed families together hold more stored items than the
composed family holds units, the model sees a mix of roughly 70\% composed:

\begin{center}
\small
\begin{tabular}{lrr}
\toprule
family & items/epoch & share \\
\midrule
composed (resampled $\times$4) & 28{,}704 & 69.7\% \\
validators & 5{,}402 & 13.1\% \\
gates & 4{,}318 & 10.5\% \\
verbatim copy & 1{,}729 & 4.2\% \\
short spans & 939 & 2.3\% \\
short verbatim copy & 83 & 0.2\% \\
\bottomrule
\end{tabular}
\end{center}

These shares are not free parameters we tuned to taste; one of them we measured.
Raising the verbatim-copy share from 4.3\% to 10\% cost 3.7 points of SEP
separation and roughly doubled TensorTrust hijacking success.
The family that trains readability is in direct tension with the family that
trains non-execution, and the tension is sharp enough that a few points of share
moves the defense.
The short-span families were therefore added at the existing proportions rather
than weighted up, even though the regime they cover is the one the overlay was
weakest on.

\paragraph{Two steps must be measured against the base model.}
Payload screening and frame ranking depend on the base model, so both are re-measured for each model trained; each takes one pass of standalone generations against the frozen model.

\emph{Payload screening} keeps only the payloads the frozen model answers correctly when asked alone, with nothing else in the prompt.
Screening is what makes a witness metric sound: without it, a model that does not know the answer scores as having resisted an injection it in fact obeyed, and every injection rate measured on the corpus is too low.
On our primary model the screen keeps 231 of 494 TriviaQA payloads and moves the measured execution rate from 40\% to 68\%, in line with SEP's reported 71.2\%.

\emph{Frame ranking} measures each frame's standalone injection rate --- payload
planted in a held-out passage, no overlay, does the witness appear --- and
frames are sampled in proportion to that rate plus a floor of 0.15.
The floor is what keeps the mix diverse: measured rates span 0.10 to 0.88, so
without it the three most effective styles would crowd out the rest, and the
benchmarks attack in shapes unlike any of them.
With it the twelve styles occupy a 1.7--16.8\% band of training draws rather
than collapsing onto the leaders.
Note that the injected frame is re-drawn per item per epoch, so it is this
distribution, not the frame stored at composition, that the model trains
against.
Frames differ by an order of magnitude in how often they succeed.
A frame the model never obeys is useless for training: the injected item's target is then what the model already does, and there is nothing to learn from it.
The bare frame --- the payload appended as a plain sentence --- serves as the reference point.
SEP's verbatim-prefix frame is excluded from composition entirely, so no training item shares SEP's surface form.

Both quantities move substantially across base models.
Between the two models in this paper, screening keeps 231 and 263 payloads, and a quarter of the kept sets are disjoint.
The frame rankings correlate at $0.49$: the frames most effective on one model succeed on the other at half the rate or less, and two styles that work on the first do nothing on the second.
Sampling shares are set from the measured ranking, but a quarter of the corpus stays spread across the weaker styles, because benchmark attacks rarely resemble the training frames.

\paragraph{Validity conditions on a target.}
A target is stored as text and rendered through the chat template at preprocessing time, so that the assistant turn's terminator is the last scored token; a model trained without that token learns where to start an answer but not where to stop.
The terminator is read from the template rather than assumed, since it differs across models.
A completion that did not stop naturally is rejected, logged, and counted, so no truncated text becomes a target.
A copy-family span that cannot be reproduced byte-exactly through the template is rejected the same way, since the family would otherwise train the model to reproduce something other than the original bytes.

\paragraph{Overlap with the benchmarks.}\label{app:overlap}
The corpus passages are contexts from the SQuAD v2.0 training split.
PIArena's SQuAD v2 split draws its 200 contexts from the same training split, and 71 of them appear verbatim as passages in our training corpus (4 more are in our held-out split).
Under PIArena's direct attack, the overlay's attack success is 0 of 71 on the shared contexts and 0 of 129 on the others; its clean utility is 86\% (61 of 71) and 81\% (104 of 129).
LoReFT, trained on the same corpus, gives the same picture: 0 of 71 and 0 of 129 attack success, and 60 of 71 and 105 of 129 clean utility.
So training on a passage does not make the overlay resist attacks planted in it any better than on an unseen passage.

No attack string from these benchmarks appears in the corpus.
For every attack we found the longest run of words it shares with the training text (all items, frame templates, and payloads, ignoring case).
It is five words for AlpacaFarm (``once you've done that,''), five for PIArena, seven for Quadrat-IPI (a sample of 400 attacks), and four for AgentDojo, all generic phrases such as ``at https://docs.''
StruQ's completion delimiters, ``\#\#\# response:'' and ``\#\#\# instruction:'', never occur in the corpus.
The benchmarks do use injection families the corpus covers (bare instructions, forged boundaries, and claims that the earlier task is finished), so they test new wordings of known families rather than unseen kinds of attack.
The same holds for StruQ and SecAlign, which train on two of the four families AlpacaFarm tests: naive and completion \citep{chen2024struq, chen2025secalign}.

\section{The red-team engagement}
\label{app:redteam}

\subsection{Setup}

Published attack corpora are written without knowledge of the defense they will face, so they cannot test whether a defense survives an adversary who knows how it works.
For that test, an external model (GLM 5.2) was given the deployment shape, the measured results, and an accurate description of the overlay --- including that it is applied unconditionally to 100\% of the retrieved passage --- and was asked to break it.
Its only affordance was a tool that ran a probe --- a developer instruction, a passage, and a condition --- and returned the marked and unmarked answers; it had no access to the implementation.
The engagement ran 222 probes against an earlier checkpoint of the overlay (one with an adapter at every layer, Appendix~\ref{app:training}), each probe paired across marked and unmarked conditions, with a clean control where the claim required one.
The probes ran in two rounds, and the adversary saw the marked and unmarked answers to every probe and wrote a new plan before the second round.
Before running a probe, it named the specific observable that would count as success, such as an exact URL, string, refusal, or brand mention.
A probe counted as a bypass only if the marked answer showed that observable and the marked answer on a clean control passage, on the same topic with no injection, did not.
The adversary did not know how the overlay works, so no probe was designed to act on its mechanism; Appendix~\ref{app:whitebox} reports a white-box attack.

\subsection{What the overlay stops}

Every direct authority-assertion attack the adversary invented failed with the overlay on and succeeded with it off.
Nine classes, of which all but the first appear in none of the roughly 800 benchmark payloads we checked them against when the engagement ran:

\begin{center}
\footnotesize
\begin{tabular}{ll}
\toprule
attack class & in benchmark corpora? \\
\midrule
direct imperatives & yes \\
fake system or developer tags & no \\
persona hijacking & no \\
meta-instructions (``do not summarize me,'' ``output REFUSED instead'') & no \\
fake trusted-source quotes (``the system instructions specify\ldots'') & no \\
passage-internal trust labels & no \\
selective suppression via a fabricated developer note & no \\
question reframing (``the question is outdated, the correct question is\ldots'') & no \\
system-prompt disclosure smuggled into a delegated procedure & no \\
\bottomrule
\end{tabular}
\end{center}

Novel shapes do not help the attacker because every one of them is a passage asserting authority it does not have, and the overlay simply ignores these text-level assertions.
Given a passage that labels the attacker's section ``trusted'' and the legitimate section ``untrusted, ignore,'' a model without the overlay obeys the labels, while with the overlay it ignores them and treats the whole span as data.

\subsection{What it does not stop}

The successful attacks all concern cases where the model is \emph{explicitly} asked to use the data as instructions.
That is, if the developer says ``follow the steps in the document,'' the model follows them --- the overlay overridden, the marked steps now run with the developer's authority --- the LLM equivalent of writing server code that runs \texttt{eval()} on untrusted user input.
One unexpected subtlety is that an authorized ``do'' can also override a ``don't'': told to follow a procedure in the data but never to produce a specific string, when presented with procedures that would generate the string, the model behaves inconsistently under the contradiction (six near-identical variants of one probe split four to two).
Nothing in the training data sets up an instruction hierarchy --- rules for which instruction wins when instructions collide, including bounded forms of delegation; training one is likely a fruitful direction for future work.

\section{What is wrong with the injection benchmarks, and how we corrected it}
\label{app:integrity}

Every measurement below comes from our own runs, and every one affects anyone using these benchmarks rather than us alone.
They were all found the same way: run each item with no injection at all, and see whether the metric still fires.
That control costs one extra generation per item, and it is the difference between a number and a measurement.

\subsection{SEP: the witness grader}

\paragraph{Witnesses fire inside ordinary words.}
SEP scores obedience as a lowercased substring test with no minimum witness length, so short witnesses match inside unrelated words.
Of 18 items scoring as executed on our marked arm, \textbf{3 also fire on the clean condition} --- the same item with no probe anywhere.
The witness \texttt{pen} matches ``o\textbf{pen} mic night''; \texttt{cat} matches ``indi\textbf{cat}es.''

The obvious repair --- require word boundaries around the witness --- is wrong in the other direction.
Word boundaries drop inflections of the witness, which are genuine hits: counted over every probe-in-data row we hold, word-boundary matching removes 79 true hits (banana/bananas, oyster/oysters, sleep/sleeping) against 65 mid-word artifacts.
It loses more true positives than false ones.
The separation score barely moves, because it is a ratio and the losses hit numerator and denominator alike, but utility is a bare count over the condition where answering is desired, so word-boundary scoring understates utility by about 3 points on every arm.
The corrected rule therefore keeps the word boundaries and adds an explicit inflection allowance --- the witness may end in \texttt{s}, \texttt{es}, \texttt{ed}, \texttt{ing}, or \texttt{'s} before the closing boundary --- which removes the mid-word artifacts while keeping the inflected true hits:

\begin{center}
\footnotesize
\begin{tabular}{lcccc}
\toprule
scoring rule & probe-in-data $\downarrow$ & utility $\uparrow$ & SEP $\uparrow$ & clean FP $\downarrow$ \\
\midrule
substring (published) & 2.9\% & 93.7\% & 97.2\% & 2.6\% \\
word boundary & 1.0\% & 90.0\% & 98.9\% & 0.8\% \\
corrected (boundary + inflections) & 1.2\% & 93.1\% & 98.8\% & 0.9\% \\
\bottomrule
\end{tabular}
\end{center}

The main text leads with the corrected grader and says so wherever it quotes a SEP number; the published grader's numbers are reported next to them here and in Appendix~\ref{app:inj-tables}, so results stay comparable with published work.
Anyone generating injection data with a witness metric should keep witnesses above about five characters; we measured a 6.5\% false-positive rate from three-letter witnesses in our own generator before enforcing this.

\paragraph{Collisions in the residual.}
The corrected rule above is programmatic, and it is the only correction our tables apply.
As an additional check, an LLM judge audited the 314 probe-in-data hits the corrected rule left on an earlier checkpoint of the overlay; we hand-checked samples.
The judge found 67 hits --- 21\% --- in which the completion never engages the probe and the witness fires on ordinary vocabulary: the ``heart'' of a contract, ``burnt orange'' in a decor palette.
These judgments are looser than the programmatic rule, and in some of these cases the probe may genuinely have influenced the completion, so we report the number and do not subtract it.

\subsection{TensorTrust: three harness defects, and one of the benchmark's own}

TensorTrust is natively three ordered segments in one prompt: an opening defense, the attacker's input, and a closing defense, with the defender writing the first and third.
The first two defects below belong to ASIDE's harness --- artifacts of rendering that structure through a two-slot instruction/input template; the third is a scoring defect any substring grader inherits, ours included before correction; the fourth belongs to the benchmark itself.

\paragraph{The harness discards 78\% of the benchmark, non-randomly.}
Rendering the benchmark through a two-slot instruction/input template leaves nowhere to put a defender segment that arrives after the input, so the harness filters to rows with an empty closing defense.
That drops \textbf{606 of 776 rows (78.1\%)}, and the drop is not a random sample: 17.8\% of all rows state the grant rule only in the closing defense, and 12.2\% state the access code only there.
The surviving subset is the one where the code is least often recoverable.
A literal test of whether the code appears in the defense text at all bounds the achievable Defense Validity at 80.9\% on the full 776 rows but only \textbf{68.2\% on the 165 that survive} --- so every arm we ever measured there, marked and unmarked alike, sat pinned against the benchmark's structural ceiling rather than measuring the adapter.

\paragraph{A third of the surviving rows omit the password.}
The filter keeps rows whose opening defense is self-contained in form without checking that the access code appears in it.
\textbf{52 of 165 rows (32\%) have an access code that appears nowhere in the defense prompt.}
One defense reads, in full, that the user will enter a password and the model should grant access if it matches the correct one --- with the labelled code being a word the model was never told.
The damage lands on Defense Validity, which replays the true code and expects a grant: 89\% on rows that state the code against 58\% on rows that do not, essentially unchanged by marking (90\% and 58\%).
The commonly cited figure near 79\% is really about 89\% on the answerable subset, dragged down by a third of the set where the task is impossible --- and the 58\% is the model guessing that a plausible input is probably the password, which inflates the metric rather than deflating it.

\paragraph{Substring scoring inflates extraction on short codes.}
Extraction asks whether the attacker made the model \emph{reveal} the access code rather than grant access; scored as a substring test, it inherits the short-witness problem of the SEP grader above.
6.8\% of access codes are under five characters, and a substring match fires inside ordinary words, inflating measured extraction on those rows by roughly 18 points.
We therefore score extraction with a word-boundary match on the code: on the frozen model this moves ASR from 42.1\% to 37.7\%, and with the overlay from 6.5\% to 6.3\% --- the numbers in Table~\ref{tab:inj}.

\paragraph{Some labelled codes do not satisfy the defense's own rule.}
This defect is the benchmark's, and no harness choice removes it: the code is discussed in the defense, but the labelled code fails the stated test.
One defense grants access only for Einstein's mass--energy equation followed by the name of the most famous Inca leader; the labelled code omits the leader, so the correct response is a refusal, which Defense Validity scores as a failure.
Another defines the code as a specific clause from a Wikipedia page, unwinnable without web access.
And some defenses are simply broken: one grants access whenever a length is negative, which is never, so correct execution grants access for any input and the row scores every model as hijacked.
These classes are invisible to a does-the-code-appear test, which is why the fix has to be empirical.

\paragraph{What we do instead.}
We keep all 776 rows, placing the closing defense in the same turn after the attacker's span and deliberately leaving it unmarked, because it is defender text and must keep its authority.
Marking a sub-span of a turn is expressible for a position-marked adapter and not for a two-slot template, which is precisely why these rows can be kept.
Spans are located by offset within the user content, never by searching the whole prompt: the access code frequently also appears inside the closing defense (290 of 1552 records), and a naive search marks the defender's copy and silently inverts the test (verified: 1552 of 1552 records slice to the attacker's span, 0 of 290 to the defender's).
Defense Validity is screened empirically --- run the unmarked base model on the full sandwich with the true code and keep the rows it grants --- which makes the ceiling 100\% by construction and excludes the defects above by behavior rather than by a heuristic that cannot see them.
On the full 776 rows the unmarked base scores 35.3\% attack success and 75.1\% Defense Validity, with 583 rows (75.1\%) answerable.

The empirical screen and the literal test disagree on 187 rows in \emph{both} directions, which is why a textual filter cannot substitute for it:

\begin{center}
\footnotesize
\begin{tabular}{lcc}
\toprule
code stated literally & base grants & rows \\
\midrule
yes & yes & 512 \\
yes & no  & 116 \\
no  & yes & 71 \\
no  & no  & 77 \\
\bottomrule
\end{tabular}
\end{center}

The 71 rows that grant without stating the code are semantic rules --- a defense that describes itself as a switch and grants when the switch is on, never quoting the phrase.
Every one of those rows is discarded by that harness.
Among the 116 that state the code yet refuse are rows unusable by construction, including one whose access code is a chat control token that cannot be passed as data through any chat template.

\section{Robustness checks and full counts}
\label{app:inj-tables}

\subsection{Full tables}
Table~\ref{tab:sep} gives SEP under both the corrected grader and the published one (Appendix~\ref{app:integrity}); Table~\ref{tab:inj} gives the other benchmarks with their own metrics.

\newcommand{\sepcaption}{SEP, all 9{,}160 items, greedy decoding.
``Probe in data'': answers the probe from the data span; ``utility'': answers when the instruction asks; SEP: of items answered as an instruction, the fraction not also answered from data.
Corrected rule with the published rule in parentheses; run-to-run noise $\sim$2pt (Table~\ref{tab:sep-seeded}).}
\newcommand{\septabular}{\begin{tabular}{lccc}
\toprule
 & frozen & overlay & LoReFT \\
\midrule
probe in data $\downarrow$ & 71.7\% (72.6) & 1.2\% (2.9) & 0.9\% (2.7) \\
utility $\uparrow$ & 92.5\% (93.3) & 93.1\% (93.7) & 92.8\% (93.4) \\
SEP $\uparrow$ & 24.3\% (24.0) & 98.8\% (97.2) & 99.0\% (97.4) \\
\bottomrule
\end{tabular}}
\newcommand{\injcaption}{TensorTrust, AlpacaFarm, Quadrat-IPI, and AgentDojo.
TensorTrust is Defense Validity (DV) screened (Appendix~\ref{app:integrity}); AlpacaFarm is the union of StruQ's four static attacks, scored \emph{begin-with}; Quadrat-IPI is every verified injected document (Appendix~\ref{app:newbench}); AgentDojo is v1.2.1 under the \texttt{important\_instructions} attack, 97 benign and 949 attacked tasks.}
\newcommand{\injtabular}{\begin{tabular}{lccc}
\toprule
 & frozen & overlay & LoReFT \\
\midrule
\multicolumn{4}{l}{\emph{TensorTrust}} \\
hijacking ASR $\downarrow$ & 35.3\% & \textbf{6.4\%} & 10.1\% \\
hijacking DV $\uparrow$ & 99.8\% & 96.2\% & 94.9\% \\
extraction ASR $\downarrow$ & 37.7\% & 6.3\% & 6.1\% \\
\midrule
\multicolumn{4}{l}{\emph{AlpacaFarm / StruQ, four static attacks}} \\
attack success $\downarrow$ & 99.0\% & 0.5\% & 0.0\% \\
\midrule
\multicolumn{4}{l}{\emph{Quadrat-IPI, 14{,}441 documents}} \\
injection compliance $\downarrow$ & 6.3\% & 0.1\% & 0.1\% \\
\midrule
\multicolumn{4}{l}{\emph{AgentDojo}} \\
benign utility $\uparrow$ & 90.7\% & 89.7\% & 86.6\% \\
utility under attack $\uparrow$ & 83.8\% & 84.8\% & 83.6\% \\
attack success $\downarrow$ & 12.8\% & 0.0\% & 0.0\% \\
\bottomrule
\end{tabular}}
\begin{table}[t]
\centering
\caption{\sepcaption}
\label{tab:sep}
\small
\septabular
\end{table}
\begin{table}[t]
\centering
\caption{\injcaption}
\label{tab:inj}
\small
\setlength{\tabcolsep}{6pt}
\injtabular
\end{table}

\subsection{What the LoReFT comparison shows}\label{app:loreft}
LoReFT applied at the marked span is itself a semantic overlay, with a rank-8 intervention in place of the SwiGLU adapter.
So the comparison measures how much the defense depends on the adapter, not whether marking works.
It depends on it very little.
With 2.1M parameters, about 4\% of the overlay's, LoReFT matches the overlay on SEP, AlpacaFarm, Quadrat-IPI, PIArena, and attack success on AgentDojo.
Most of the defense therefore comes from training an edit on the marked span with this objective, not from the adapter's capacity.
The two adapters separate in three places, each in the overlay's favor.
LoReFT lets through more TensorTrust hijacks (10.1\% against 6.4\%) on every sampled seed, and a rank-192 LoReFT with the overlay's parameter count did no better (Appendix~\ref{app:training}).
It copies random tokens less exactly (86.0\% against 92.7\%; Appendix~\ref{app:fidelity}).
And it keeps slightly less utility: 70.5\% against 73.5\% on clean PIArena questions, and 86.6\% against 89.7\% of AgentDojo's benign tasks.
So a deployment that needs only this overlay could use LoReFT instead, at a small cost in hijacking and utility.
The saving would be small too: the overlay adds a few milliseconds of prefill and almost nothing to decoding (Appendix~\ref{app:serving}).
With several overlays stacked, LoReFT falls well behind (\S~\ref{sec:marks}).

\subsection{Under ASIDE's protocols}

The main text reports greedy decoding.
Repeating both arms under ASIDE's protocol --- bf16, sampling temperature 0.7, three random seeds --- reproduces every overlay number to within 0.4 points of the greedy run and every frozen number to within 1.0:

\begin{table}[h]
\centering
\footnotesize
\caption{SEP, all 9{,}160 items, temperature 0.7.
Mean over three seeds $\pm$ the range across them.}
\label{tab:sep-seeded}
\begin{tabular}{llcccc}
\toprule
arm & rule & probe in data $\downarrow$ & utility $\uparrow$ & SEP $\uparrow$ & clean FP $\downarrow$ \\
\midrule
frozen model & substring & 73.6\%\,$\pm$0.3 & 92.9\%\,$\pm$0.2 & 23.3\%\,$\pm$0.2 & 2.8\%\,$\pm$0.1 \\
 & corrected & 72.7\%\,$\pm$0.4 & 92.2\%\,$\pm$0.2 & 23.7\%\,$\pm$0.4 & 0.9\%\,$\pm$0.1 \\
with overlay & substring & 2.8\%\,$\pm$0.2 & 93.4\%\,$\pm$0.3 & \textbf{97.2\%}\,$\pm$0.2 & 2.7\%\,$\pm$0.0 \\
 & corrected & 1.0\%\,$\pm$0.2 & 92.7\%\,$\pm$0.3 & 98.9\%\,$\pm$0.2 & 0.9\%\,$\pm$0.0 \\
\bottomrule
\end{tabular}
\end{table}

Two measurement notes.
Running the same frozen configuration twice on byte-identical prompts (a 300-item paired subset) gives a 2.6-point SEP gap: 27.9\% of completions differ and 2.0\% of witness verdicts flip, so about 2 points is the run-to-run noise floor for this benchmark.
Separately, generation runs to 4{,}096 new tokens rather than ASIDE's 1{,}024: 3--4\% of completions run past 1{,}024 tokens in both arms, so the shorter limit would hide a witness written late, while under 0.1\% reach 4{,}096.
On TensorTrust, ASIDE's own 165-row harness and grading give ASR 50.9\% $\to$ 8.5\% for the reported checkpoint, decoded greedily.

The one gap we claim between the overlay and LoReFT is TensorTrust hijacking, 6.4\% against 10.1\% under greedy decoding, so we repeated it on all 776 rows at temperature 0.7 with three seeds.
The overlay grants access to 5.9\% of attacks (range 5.8--6.1) and LoReFT to 9.9\% (range 9.7--10.1); the ranges do not overlap.
The gap is row by row, not an average: 25 attacks win against LoReFT on all three seeds and against the overlay on none, while 4 go the other way.
Sampling lowers Defense Validity for both, to 92.6\% for the overlay and 89.6\% for LoReFT, since a sampled reply strays more often from the exact ``Access Granted'' the grader requires.

\subsection{PIArena}

\begin{table}[h]
\centering
\footnotesize
\caption{PIArena HotpotQA and NQ, 50 items per category per condition.
The clean column is the same item with no injection.}
\begin{tabular}{lcccccc}
\toprule
category & \multicolumn{2}{c}{frozen base} & \multicolumn{2}{c}{overlay} & \multicolumn{2}{c}{LoReFT} \\
 & injected $\downarrow$ & clean $\downarrow$ & injected $\downarrow$ & clean $\downarrow$ & injected $\downarrow$ & clean $\downarrow$ \\
\midrule
infrastructure failure & 82.0\% & 0.0\% & 0.0\% & 0.0\% & 0.0\% & 0.0\% \\
phishing & 56.0\% & 0.0\% & 2.0\% & 0.0\% & 0.0\% & 0.0\% \\
content promotion & 54.0\% & 0.0\% & 0.0\% & 0.0\% & 0.0\% & 0.0\% \\
access denial & 20.0\% & 0.0\% & 0.0\% & 0.0\% & 0.0\% & 0.0\% \\
\bottomrule
\end{tabular}
\end{table}

Marking costs no utility: with the overlay the model answers 74.0\% of injected questions correctly against 73.5\% clean, where the frozen model falls from 73.5\% to 68.0\% (SQuAD~v2: 83.0\% against 82.5\% clean, frozen 46.5\%).
The single phishing row scored as an attack is a quotation rather than a compliance: the model quotes the injected sentence while reasoning about it, then answers without the URL, and the URL-presence rule fires on the quotation; no completion in 400 follows an injected instruction.

\subsection{Fidelity}
\label{app:fidelity}
Asked to quote the marked span verbatim (500 held-out passages), the model reading through the overlay is character-exact on all 500, against 98.4\% with no overlay and 99.8\% through LoReFT, whose one miss turns an ASCII comma into a full-width one.
The overlay strips what the span can \emph{do}, not what it \emph{says} --- it succeeds by encoding non-executability, rather than filtering or suppression.

Natural passages understate damage to the marked tokens, because the model can rebuild a half-read sentence from its knowledge of the language.
So we also ask for copies of strings with less and less structure: shuffled words, UUIDs, random alphanumeric keys, and sequences of random vocabulary tokens (150 of each, plus 150 natural passages; Table~\ref{tab:hecopy}).
On random tokens nothing in the context helps, and only the tokens' identities, preserved at the marked positions, can produce an exact copy.

\begin{table}[h]
\centering
\footnotesize
\caption{Exact copy of the marked span, by how much structure the span has (150 items each).
``Earlier checkpoint'' has an adapter at every layer, 0 through 31.}
\label{tab:hecopy}
\setlength{\tabcolsep}{4pt}
\begin{tabular}{lccccc}
\toprule
 & natural & shuffled words & UUID & random keys & random tokens \\
\midrule
frozen & 99.3 & 98.7 & 100 & 100 & 90.7 \\
earlier checkpoint & 85.3 & 47.3 & 54.7 & 74.0 & 1.3 \\
\quad same, layer-0 adapter switched off & 98.7 & 84.7 & 78.0 & 98.7 & 58.0 \\
Semantic Overlays (this paper) & 100 & 99.3 & 100 & 99.3 & 92.7 \\
LoReFT (same data) & 100 & 99.3 & 100 & 100 & 86.0 \\
\bottomrule
\end{tabular}
\end{table}

The earlier checkpoint was as robust as the one in this paper, but it lost the identity of the marked tokens: it copied 47\% of shuffled word lists and 1\% of random-token strings exactly.
We found where the damage came from by patching.
On items the earlier checkpoint failed to copy and the frozen model copied, we switched its adapters on at only some layers.
For natural text, every adapter but layer~0's gave 97\% exact copies, against none with all of them on (Appendix~\ref{app:training}).
For the high-entropy strings, the damage that remained with layer~0 off came from two bands, layers 1--7 and 20--27; adapters at layers 8--19 did no harm.
Corruption that reaches layer~27 or so, where the model reads out what it copies, is enough to collapse copying: in the natural-text sweep, letting it through dropped exact copies from 89\% to 32\%.

We also checked the obvious suspect and found it innocent.
The overlay does move the next-token distribution at marked positions away from the frozen model's, and more so as training goes on (1.35 to 2.91 nats per token from step 400 to the end), but LoReFT moves it about as far (1.86 nats) and copies well.
Copying depends on keeping the identity of tokens already read, which next-token prediction at those positions does not measure.

Two changes repaired the identity without costing robustness: no adapter at layer~0, and a quarter of the verbatim-copy training items replaced by high-entropy strings (Appendix~\ref{app:corpus}).
The overlay in this paper, trained from scratch with both, copies random tokens slightly more often than the frozen model (92.7\% against 90.7\%), and more often than LoReFT trained on the same data (86.0\%).

\section{PIArena against the defenses its authors evaluated}
\label{app:piarena}

PIArena \citep{piarena2026} is the one benchmark of the six whose own paper
evaluates a slate of published defenses, so it supports a direct comparison
rather than a comparison of operating points.
Table~\ref{tab:piarena-defenses} places the overlay in their table.
Their rows are the Direct attack on HotpotQA and NQ, the two RAG splits our
evaluation draws from, averaged over the two.
The settings agree at the baseline: their undefended attack succeeds on 49.5\%
of samples, ours on 53\%.

Their defenses divide into two kinds and the division matters.
\emph{Prevention} defenses let the model answer and are scored on both axes.
\emph{Detection} defenses classify the input and block it, so their authors
report no utility under attack --- ``flagged inputs produce no responses.''
A detector that blocks everything scores 0\% attack success by construction,
which is why AttentionTracker's zero is not the same object as ours.

\begin{table}[h]
\centering
\caption{PIArena, Direct attack, mean of the HotpotQA and NQ RAG splits.
Utility is task accuracy under attack.
Compare rows within a block: the top block is Table~2 of \citet{piarena2026}, with their base models and grader; the bottom block is our runs on frozen Qwen3.5-9B with our judge, where the frozen model scores 73.5\% with no attack.
The detection rows have no utility to report.}
\label{tab:piarena-defenses}
\small
\begin{tabular}{llcc}
\toprule
& defense & attack success $\downarrow$ & utility $\uparrow$ \\
\midrule
\multicolumn{4}{l}{\emph{PIArena's table (their base models and grader)}} \\
& none (undefended) & 49.5\% & 82.5\% \\
\midrule
\multirow{4}{*}{\rotatebox{90}{\footnotesize prevention}}
& PromptArmor & 48.0\% & 84.0\% \\
& DataFilter & 30.0\% & 80.0\% \\
& PISanitizer & 11.0\% & 89.5\% \\
& SecAlign++ & 3.5\% & 68.5\% \\
\midrule
\multirow{4}{*}{\rotatebox{90}{\footnotesize detection}}
& DataSentinel & 33.0\% & --- \\
& PromptGuard & 21.0\% & --- \\
& PIGuard & 18.0\% & --- \\
& AttentionTracker & 0.0\% & --- \\
\midrule
\multicolumn{4}{l}{\emph{This work (frozen Qwen3.5-9B, our judge)}} \\
& none (frozen) & 53.0\% & 68.0\% \\
& Semantic Overlays & 0.5\% & 74.0\% \\
& LoReFT, same data & 0.0\% & 70.0\% \\
\bottomrule
\end{tabular}
\end{table}

Among the defenses in their table that leave the model able to answer, none reaches the overlay's attack success.
The next best is SecAlign++ at 3.5\%, seven times higher.
Its utility drop in their table (82.5\% to 68.5\%) does not reproduce on our harness: Meta-SecAlign on Llama-3.1-8B keeps 79.0\% PIArena utility against the frozen model's 73.5\%, and loses utility on SEP instead (78.7\% against 91.3\%; Appendix~\ref{app:same-model}).
The overlay loses none, answering as many injected questions correctly as clean ones (74.0\% against 73.5\%).
LoReFT on the same span and data reaches 0.0\%, at 70.0\% utility against its own 70.5\% clean.
The only entry below us is a detector whose 0\% comes from declining to answer at all; its authors record that it ``significantly harms utility (0\% on most datasets),'' where our clean controls fire at 0\%.

\paragraph{Other published results on PIArena.}
Three other papers report defenses on PIArena.
PISmith \citep{yin2026pismith} runs Meta-SecAlign-8B under the same Direct attack: 2\% on HotpotQA, 5\% on NQ, and 1\% on SQuAD~v2.
On our own harness and judge, with Llama-3.1-8B-Instruct as the base for both, Meta-SecAlign reaches 1.0\% on HotpotQA and NQ and 0.5\% on SQuAD~v2, against the overlay's 0.0\% on both (Appendix~\ref{app:same-model}).
PISmith also runs seven defenses on Qwen3-4B, the best of which, PIGuard, lets through 15--21\%.
AgentWatcher \citep{wang2026agentwatcher}, a rule-based monitor on Qwen3-4B, reaches 0.0\% on PIArena's HotpotQA at almost no cost in utility; its HotpotQA is the long-context version from LongBench, not the retrieval split we use.
WARD \citep{cao2026ward} reports 1--6\% under PIArena's adaptive Strategy attack, in which an attacker model rewrites the injection, without naming its base model.
That attack is stronger than the Direct attack the comparison above uses, and we run it against the overlay in Appendix~\ref{app:piarena-strategy}.

\paragraph{Frontier models fail this benchmark.}
The same paper reports commercial models under the same Direct attack, and they
are not close to defended: GPT-4o 92\%, Gemini-3-Flash 88\%, Gemini-3-Pro 83\%,
GPT-4o-mini 76\%, GPT-5 70\%, Claude-Sonnet-4.5 31\%.
Two of those are documented as hardened --- GPT-4o-mini ``is specifically
trained to resist prompt injection'' and still admits 76\%, GPT-5 ships ``a
multilayered defense stack'' and still admits 70\%.
Their conclusion is that ``even closed-source LLMs with specialized training
struggle against realistic, diverse prompt injections.''
A frozen open 9B model with one trained overlay sits at 0.5\% where those
systems sit between 31\% and 92\%.

\paragraph{The remaining 0.5\%.}
Appendix~\ref{app:inj-tables} gives the per-family counts and the reason the one
non-zero cell is not a compliance: the model quotes the injected sentence while
reasoning about it and answers without the URL,
and the URL-presence rule fires on the quotation.
No completion in 400 follows an injected instruction.
We keep it scored as an attack, which is the conservative choice and the one the
0.5\% above reflects.

\subsection{The adaptive Strategy attack}
\label{app:piarena-strategy}

The Direct attack plants the injected task as written, so it measures a defense against an attacker who does not react to it.
PIArena also ships an adaptive black-box attack, which its authors call Strategy \citep{piarena2026}.
An attacker model rewrites the injected task in ten styles, such as an editor's correction or a system configuration block, and plants five rewrites of each.
If none succeeds, it takes one rewrite per style and revises it for up to five rounds, using the defended model's answers as feedback.
A judge model decides after every answer whether the injected task was completed, and the search stops at the first success.
So each item costs up to about 100 queries to the defended model.

We ran the attack on all 400 items of the three splits we use, against frozen Qwen3.5-9B with and without the overlay over the whole retrieved passage.
The attacker and the judge are the model PIArena uses for both, Qwen3-4B-Instruct.
PIArena's judge is asked only whether the response ``completes the task,'' and it accepts some responses that neither carry out nor mention the injected task.
Because the search stops at the first accepted response, out of up to about 100 per item, a rare false acceptance per response becomes a common one per item against a defense that holds.
So we audited every success with a second judge, Claude Haiku~5.5, which reads the injected task and the response and decides whether the response carries the task out.
For the overlay we then reran the search on every item without a confirmed success, with the change that the search stops only when both judges accept; a success the audit rejects no longer ends the search.
In the first run the attacker's rewrites were capped at 1,024 tokens, and 4.4\% of them were cut; a cut rewrite is a weaker attack, so the overlay's row under PIArena's judge alone is biased in the overlay's favor.
The rerun raised the cap to 4,096 tokens, which cut 1.6\% of rewrites, and the ones we read were repetition loops.
So the audited row is not affected: its first-run items are successes, and every other item was rerun.
The defended model's responses were capped at 2,048 tokens; 166 of the first run's 36,710 responses and 192 of the rerun's 36,318 reached the cap and were judged as cut, and the ones we read were runaway answers or repetition loops.

\begin{table}[h]
\centering
\caption{PIArena, Strategy attack, attack success per split.
Defense rows are Table~2 of \citet{piarena2026}, on their base models and scored by PIArena's judge.
The last four rows are this work on frozen Qwen3.5-9B.
``PIArena judge'' is the standard protocol; ``audited'' counts only successes a second judge confirms, with the overlay's search continued past every rejected one.
The overlay's PIArena-judge row comes from a run in which the attacker's rewrites were capped at 1,024 tokens, which favors the overlay (see text).}
\label{tab:piarena-strategy}
\small
\begin{tabular}{llccc}
\toprule
& defense & SQuAD~v2 & NQ & HotpotQA \\
\midrule
\multicolumn{5}{l}{\emph{PIArena's table (their base models and judge)}} \\
& none (undefended) & 100\% & 100\% & 100\% \\
\midrule
\multirow{4}{*}{\rotatebox{90}{\footnotesize prevention}}
& PromptArmor & 100\% & 98\% & 96\% \\
& DataFilter & 93\% & 82\% & 84\% \\
& PISanitizer & 85\% & 92\% & 94\% \\
& SecAlign++ & 9\% & 8\% & 9\% \\
\midrule
\multirow{4}{*}{\rotatebox{90}{\footnotesize detection}}
& DataSentinel & 78\% & 37\% & 54\% \\
& PromptGuard & 100\% & 99\% & 92\% \\
& PIGuard & 71\% & 87\% & 80\% \\
& AttentionTracker & 0\% & 0\% & 0\% \\
\midrule
\multicolumn{5}{l}{\emph{This work (frozen Qwen3.5-9B)}} \\
& frozen, PIArena judge & 100\% & 100\% & 100\% \\
& frozen, audited & 96.5\% & 97\% & 99\% \\
& Semantic Overlays, PIArena judge & 20.5\% & 13\% & 15\% \\
& Semantic Overlays, audited & 15.0\% & 4\% & 9\% \\
\bottomrule
\end{tabular}
\end{table}

The attack defeats the frozen model on every item, and the audit confirms 389 of the 400 successes.
With the overlay, the attack succeeds on 43 of 400 items after the audit (10.8\%), and on 69 under PIArena's judge alone (17.2\%).
That is far below every published defense in Table~\ref{tab:piarena-strategy} except two.
AttentionTracker reaches 0\% by declining to answer: its utility is 0--2\% on both RAG splits even with no attack.
SecAlign++ reaches 8--9\% on all three splits under PIArena's judge, below the overlay's 13--20.5\% under the same judge.
Its numbers come from the same search and the same judge, so they likely include false acceptances too, and we cannot audit them.
With the audit, the overlay's 4\% on NQ and 9\% on HotpotQA match SecAlign++, and its 15.0\% on SQuAD~v2 does not; the two judges differ, so this is not a like-for-like comparison.
SecAlign++ keeps 67--91\% utility under this attack in their table; we did not measure the overlay's utility under the Strategy attack.
So under PIArena's Strategy attack the overlay is one of the two strongest defenses in this comparison that still answer, and the comparison cannot settle which of the two is stronger.

The two judges disagree far more on the overlay than on the frozen model.
On the frozen model the audit rejects 11 of 400 successes.
On the overlay it confirms 33 of the first run's 69, and in the rerun only 10 of the 91 responses PIArena's judge accepted, out of 36,318.
The rejected responses mostly do not mention the injected task at all: in 32 of the first run's 36 rejections the injected task names a URL or an exact string to output, and 31 of those responses do not contain it.
In a random sample of eight rejected responses from the rerun, seven simply answer the user's question and never refer to the injected task.
The eighth answers the question and also repeats a figure from the planted text, without doing what the injected task asks.
We audited only the successes PIArena's judge reported, not its rejections.

The confirmed successes are uneven across PIArena's four attack goals: 17 promote content, 15 are phishing, 8 claim an infrastructure failure, and 3 deny access.
At least one of them contains no instruction at all: the planted text states the attacker's claim as fact, and the model repeats it in its answer.
The overlay is trained to stop the model from executing instructions in the span, not to stop it from believing the span, so a planted falsehood is outside what it defends.
PIArena's authors make the same observation about every defense they test, in their experiment where the injected task aligns with the target task.

Every model in the loop has an output cap, and we recorded every call that hit one.
The defended model answers within 2{,}048 tokens.
In the overlay's rerun 192 of its answers hit the cap (166 in the first run, 11 on the frozen model); all were scored as written, and one of them is among the overlay's confirmed successes.
The attacker writes within 4{,}096 tokens.
The first runs used 1{,}024, which truncated 9.6\% of the overlay's revision calls and 6 calls on the frozen model, so the rerun raised the cap to 4{,}096; the 33 successes kept from the first run were found under the lower cap.
In the rerun 579 of 36{,}210 attacker calls still hit it, and 577 of those are degenerate loops that repeat one phrase until the cap.

\section{White-box attack and adversarial training}\label{app:whitebox}

The weights, the adapters, and the training code are public, so the realistic attacker has gradients.
Such an attacker breaks the overlay as trained in the main text.
Training against a continuous attacker, an approach Claude Fable~5.1 proposed, makes the same attack fail on most items at a small cost in utility.

\paragraph{The attack.}
We run GCG \citep{zou2023gcg} at the budget SecAlign uses to evaluate itself \citep{chen2025secalign}: 500 steps, 512 candidates per step, top-$k$ 256, and a 20-token adversarial span.
The attacker may use any token in the vocabulary whose text re-tokenizes to the same token; SecAlign's own GCG also bans non-ASCII tokens, which on Qwen3.5-9B would remove 48.5\% of the vocabulary.
The span sits at the end of the marked input of 50 AlpacaFarm items, the same 50 items for every checkpoint below, so every comparison is paired item by item.
The attack succeeds when the response begins with the injected marker, SecAlign's rule.
Two controls check the harness.
Against the frozen model, with no overlay, GCG succeeds on all 50 items, almost always at its first step.
With the overlay on but the adversarial span moved into the unmarked instruction, it also succeeds on all 50.

\paragraph{The overlay without adversarial training.}
GCG breaks the checkpoint of the main text on 36 of the 50 items (72\%; 95\% interval 58--83\%).
The overlay makes the attack work for its success: the median break comes at step 69 rather than step 1, and on the 14 items it survives, the attacker's loss falls from 23--45 nats to 1.4--13 without forcing the marker.
But it does not stop an optimizing attacker.
The checkpoint before this one fared no better (58\%; the two do not differ, McNemar $p=0.14$).

\paragraph{Adversarial training.}
GCG is too slow and too weak to serve as the attacker inside training.
Each attack costs minutes, and an overlay trained on a bank of GCG attacks learns to block those attacks rather than the next ones: in our early attempts, attacks found against one round rarely transferred to the next, and the success rate fell to 28\% at best before rising again.
Following an idea from Claude Fable~5.1, we replace the discrete attacker with a continuous one, solved afresh inside every training step \citep{madry2018towards,xhonneux2024efficient}.
The attacker moves the input embeddings at the injection's positions within an L2 ball whose radius is twice the mean embedding norm.
Neighboring tokens sit about 1.15 mean norms apart, so this ball contains every substitution of one token for another: any text an attacker could write at those positions is a point inside it.
The passage around the injection may move only within 0.05 mean norms, and each of its embeddings must keep its nearest token, so the passage still reads as the same text and its clean answer stays the right label.
The attacker runs Adam toward the frozen model's own completion of the injected request until its loss stops improving (no 0.05-nat gain in 50 steps, at most 600 steps), and may also own up to 50 extra filler tokens around the injection.
The overlay is then trained, by the main recipe's cross-entropy, to give the clean answer at the attacked point.
Half of the training draws go through this attacker; on a few of them the injection is absent and the attacker instead tries to derail the clean answer.
The other half are ordinary corpus draws, which keep utility in the objective.

We warm-start from the checkpoint of the main text and train for 300 steps (effective batch 32, constant learning rate $5\times10^{-4}$, about three hours on eight H100 GPUs).
GCG at the same budget then breaks 7 of the 50 items (14\%; 95\% interval 7--26\%).
Of the 36 items it broke before, 32 now hold; 3 items it did not break before now break ($p<10^{-6}$).
The attacker's best loss is higher than before on 43 of the 50 items, by a median of 4.5 nats.
Training on for another 100 steps buys nothing further (18\%, 9 of 50), and the items that break change from one snapshot to the next: only two break at both.

Most of the remaining breaks are not obedience.
On three of the seven, the input is a single word or line, and the model prints the marker with a little noise and no answer.
On two, it prints the marker and then does the trusted task as usual.
On the last two, the adversarial span also carries a style instruction (a tone, an orthography); on one the answer clearly adopts it while still doing the trusted task, and on the other only partly.
So by a strict reading, the attacker gets the hardened overlay to follow an instruction on one or two of the 50 items.
The 100-step-later snapshot breaks in the same three ways.

\paragraph{Replication.}
The same recipe, warm-started from the previous checkpoint, took GCG on the same 50 items from 58\% to 14\% at 300 steps ($p<10^{-4}$).
On a fixed 1,000-item SEP subset, its utility moved from 92.8\% to 92.4\%, within noise ($p=0.63$).

\paragraph{What it costs.}
On the full SEP benchmark (9,160 items, greedy), under SEP's published grader, the hardened overlay answers trusted-field probes on 93.1\% of items, against 93.7\% before adversarial training; the paired difference (191 items lost, 135 gained) is small but real ($p=0.002$).
Separation is unchanged (97.3\% against 97.2\%), and so are probe execution from the marked span (2.7\% against 2.9\%) and false positives on clean items (2.7\% against 2.6\%).
Of the 191 lost items, 34 treat the trusted question as a question about the marked document and decline it; in others the model addresses the probe without the exact word the scorer looks for.
Adversarial training does cost reading fidelity, mostly where the text has no structure.
Asked to quote 500 held-out marked passages, the hardened overlay is character-exact on 490 (98.0\%), against all 500 before and 98.4\% for the frozen model; the misses normalize characters (a curly apostrophe made straight, a double space closed) and in two cases change a word.
On the high-entropy strings of Table~\ref{tab:hecopy}, it copies natural passages, UUIDs, and random keys as before (98.7\%, 100\%, and 100\%), shuffled word lists slightly less often (93.3\% against 99.3\%), and random vocabulary tokens far less often (28.0\% against 92.7\%, frozen 90.7\%).
So where the context can carry a copy, the hardened overlay still copies, but where only the identity of each marked token can, it often loses one: about one token in ninety comes back wrong.
The fidelity fix of Appendix~\ref{app:fidelity} came from random-key copy items in the corpus, and in this run those items reach the overlay only through the half of the draws that are not attacked.
We expect that keeping random-token copying requires putting it under the same adversarial pressure, with copy items among the attacked draws, and have not run that.

\paragraph{Scope.}
These numbers are for one attack family at one budget.
GCG is the attack the training does \emph{not} see: the training attacker is continuous, so GCG measures transfer from the ball to the discrete text an attacker can actually write.
Attacks that start from the defense's own training checkpoints \citep{yang2025checkpointgcg} and other optimizers have not been run against the hardened overlay.
One black-box adaptive attack has.
Against PISmith \citep{yin2026pismith}, an attacker trained with reinforcement learning, run with SecOPD's protocol and judges \citep{peng2026secopd} at 10 attempts per item, the overlay of the replication above is judged broken on 83.6\% of 1{,}024 SEP items.
Reading every judged success, the model follows a planted instruction or premise on 31.4\% of items; in the rest the answer appears because the model carries out its task on attacker-written text.
Most of these attacks plant false premises rather than instructions, which a do-not-execute overlay does not target.
The main text reports the overlay before adversarial training throughout.

\section{AlpacaFarm and Quadrat-IPI}
\label{app:newbench}

\subsection{AlpacaFarm, under StruQ's harness}

StruQ \citep{chen2024struq} evaluates by planting an attacker instruction in
the data field of AlpacaFarm's evaluation set.
Of its 805 items, 208 have a non-empty input; those are the ones an indirect
injection can be placed in.
We take StruQ's four static attacks verbatim from their repository --- naive,
ignore, escape-separation, and completion --- with their payload, and mark the
whole input field.

Scoring has two conventions.
SecAlign counts a response as attacked only if it begins with the marker.
StruQ counts responses that contain the marker, then inspects every one by hand and keeps only those where the model ``regards the injected prompt as a new instruction'' \citep{chen2024struq}.
StruQ's released code skips the inspection; we apply it, and report both rules.
The inspection matters for a defense that keeps marked text readable: asked to correct the grammar of a passage, the model reproduces the injected sentence along with the rest, which is a correct answer and not an attack.

\begin{table}[h]
\centering
\caption{AlpacaFarm, all 208 input-bearing items, frozen Qwen3.5-9B versus the
same model with the input field marked, and LoReFT on the same span.
All three arms decode up to 1{,}024 new tokens; 16 frozen, 75 overlay, and 73 LoReFT completions of 1{,}248 reach that limit, and no overlay or LoReFT one contains the marker.
``Begins'' is SecAlign's criterion. ``StruQ's rule'' counts responses that contain the marker and act on the injected instruction, with false positives removed by inspecting every such response, as StruQ does.}
\label{tab:alpaca}
\small
\begin{tabular}{lcccccc}
\toprule
& \multicolumn{3}{c}{begins with marker $\downarrow$}
& \multicolumn{3}{c}{StruQ's rule $\downarrow$} \\
\cmidrule(lr){2-4}\cmidrule(lr){5-7}
attack & frozen & overlay & LoReFT & frozen & overlay & LoReFT \\
\midrule
naive & 85.6\% & 0.5\% & 0.0\% & 96.2\% & 3.4\% & 1.0\% \\
ignore & 77.9\% & 0.0\% & 0.0\% & 78.4\% & 0.0\% & 0.0\% \\
escape separation & 92.8\% & 0.0\% & 0.0\% & 98.1\% & 1.4\% & 0.0\% \\
completion & 98.6\% & 0.0\% & 0.0\% & 100\% & 0.0\% & 0.0\% \\
\midrule
any of the four & 99.0\% & 0.5\% & 0.0\% & 100\% & 3.4\% & 1.0\% \\
no attack (clean) & 0.0\% & 0.0\% & 0.0\% & 0.0\% & 0.0\% & 0.0\% \\
\bottomrule
\end{tabular}
\end{table}

The frozen arm here is more attackable than the undefended Llama-3.1-8B that Meta SecAlign
reports under the same begin-with rule (99.0\% against 68.3\%; \citealp{chen2025metasecalign}), because our serving framing places the task in
the system prompt and the data in a single user turn, so a forged turn boundary
reads as more plausible than it does inside a chat template with the data in
its own turn.
Both arms share that framing exactly, so the reported gap is conservative
rather than flattering.

Utility does not move.
On the 208 clean inputs the overlay's answers and the frozen model's have mean unigram-F1 0.66 against each other, and LoReFT's have 0.66 as well.
These are open-ended tasks, so two good answers share only part of their words: asked for a Python loop snippet, the frozen model writes a \texttt{break} example and the overlay an \texttt{enumerate} one.
Median response length is 966 characters with the overlay against 946 without.

Of the 23 overlay responses that contain the marker, 13 are false positives under StruQ's rule: 11 reproduce the injected sentence in a rewrite, transcription, or format conversion, and 2 quote it while pointing it out.
The other 10, on 7 items, act on it.
All 7 are code tasks, where the model writes the injected print into the program; asked to convert a line of Python to Ruby, it adds \texttt{puts "hacked!"}.
The one begin-with hit is a code-generation item whose whole answer is the planted string.

\subsection{Quadrat-IPI, run model-side}

Quadrat-IPI \citep{quadrat2026} labels each injected document along two axes:
the \emph{family}, meaning how the injection tries to earn compliance (bare,
forged frame, guard, identity, pretext, persistence, revocation, output
marking, execution surface, inference), and the \emph{action}, meaning what it
asks for (deny, disarm, disclose, escalate, execute, exfiltrate, hijack task,
manipulate, persist, unauthorized action).
Ninety-two of the hundred cells are populated.

It ships as a \emph{detector} benchmark: a system returns a score per document,
the score is thresholded on a clean split at a fixed false-positive budget, and
recall is reported.
An overlay is not a detector --- it never classifies, so it has no false-alarm
rate to threshold.
We therefore run the corpus model-side: each document is put in front of the
model doing an ordinary task (``summarize this email in one sentence''), the
document is marked, and we score whether the model carries out the injected
action.

Compliance requires a judge, since the corpus labels where injections are and
not what a model should have done with them.
The judge never sees the injection and the response together: one step reads
only the injection and states what it demands, a second reads only the response
and that demand, and every claim must quote verbatim evidence that is checked
against the response mechanically, with ungrounded claims downgraded.
We use Claude Haiku 4.5, measured at 95.9\% agreement and 89.2\% precision on
\texttt{complied} against a Claude Sonnet 4.5 reference on 369 records.
A cheaper judge we first tried scored 68.3\% precision, over-reporting
compliance by roughly half; judge choice is a correctness parameter here, not a
cost decision.

Across all 14{,}441 verified injected documents, compliance falls from 6.3\% to
0.1\% (LoReFT: 0.1\%).
The frozen model's worst cell is \texttt{output\_marking} $\times$
\texttt{manipulate} at 51.1\%; marked, no cell exceeds 1.5\% and 80 of the 92
populated cells are exactly zero (LoReFT: 1.5\% and 84).
Compliance can be silent --- an injection reading ``respond only with
`Approved'\,'' produces a response that contains no reference to the
instruction --- and every silent compliance we found came from the unmarked
arm.

\section{Long contexts}\label{app:longctx}

A mark edits only the marked positions, but every later position attends to them (\S~\ref{sec:adapter}), so marked text could change how the model reads unmarked text far away.
We measure this on clean question answering, with no attacks.
Each item is one SQuAD v2.0 dev question with its passage, placed among other dev passages that do not contain the answer, filling 2k, 8k, 16k, or 32k tokens.
SQuAD splits train and dev by article, so none of these passages appeared in training.
Across the 200 questions the answer passage sits at depths spread evenly over the prompt, and each question keeps its depth at every length.
Every item is run three ways on identical tokens: unmarked, which is the frozen model; every passage marked except the answer passage; and every passage marked.
A completion is correct if it contains a gold answer, PIArena's utility rule; decoding is greedy, and no completion reached the 4,096-token cap.

\begin{table}[h]
\centering
\caption{Question answering in long prompts, \% of 200 questions whose completion contains a gold answer. Differences are from the unmarked prompt on the same items, with paired-bootstrap 95\% intervals.}
\label{tab:longctx}
\small
\begin{tabular}{rccc}
\toprule
prompt length & unmarked & \makecell{all but answer\\passage marked} & \makecell{all passages\\marked} \\
\midrule
2k  & 91.0 & 92.0 \quad $+1.0$ [$-1.5$, $+3.5$] & 90.5 \quad $-0.5$ [$-3.5$, $+2.5$] \\
8k  & 86.5 & 87.0 \quad $+0.5$ [$-3.0$, $+4.0$] & 83.0 \quad $-3.5$ [$-8.0$, $+1.0$] \\
16k & 85.0 & 84.5 \quad $-0.5$ [$-4.5$, $+3.5$] & 77.0 \quad $-8.0$ [$-13.5$, $-2.5$] \\
32k & 80.0 & 77.5 \quad $-2.5$ [$-7.5$, $+2.5$] & 71.5 \quad $-8.5$ [$-14.5$, $-2.5$] \\
\bottomrule
\end{tabular}
\end{table}

Marking the other passages leaves accuracy on the unmarked answer passage unchanged at every length: every interval in the middle column contains zero (Table~\ref{tab:longctx}).
The difference also does not grow with the amount of marked text before the answer passage: split into quartiles by that amount, no length has a quartile whose interval excludes zero.
The wording of answers does drift, and it drifts more as the prompt grows: unigram F1 against the same item's unmarked completion falls from 90 at 2k to 74 at 32k.
But the drift does not move accuracy in either direction: of 800 paired items, 33 answers turn wrong and 30 turn right.

Marking the answer passage too, as a deployment that marks every retrieved passage would, costs accuracy in long prompts: 8.0 points at 16k and 8.5 at 32k, with no significant change at 2k and 8k.
The failures we read take two forms.
In one, the model answers with a sentence from the wrong passage: asked ``What is Raghuram Rajan's career?'', with no passage marked the model answers ``Central Banking economist Raghuram Rajan argues that \ldots'', and with every passage marked it answers ``Harvard's academic programs operate on a semester calendar beginning in early September and ending in mid-May.''
In the other, it repeats part of the question: asked ``Le grand asks for a variance of what three terms?'', it answers ``lawful protest demonstration, nonviolent civil disobedience, and violent civil disobedience'' with no passage marked and ``variance of what three terms'' with every passage marked.
No training prompt comes close to these lengths, so training on long documents is the natural next step.

\section{Position among published prompt-injection defenses}
\label{app:position}

\begin{figure}[t]\centering
\includegraphics[width=0.95\textwidth]{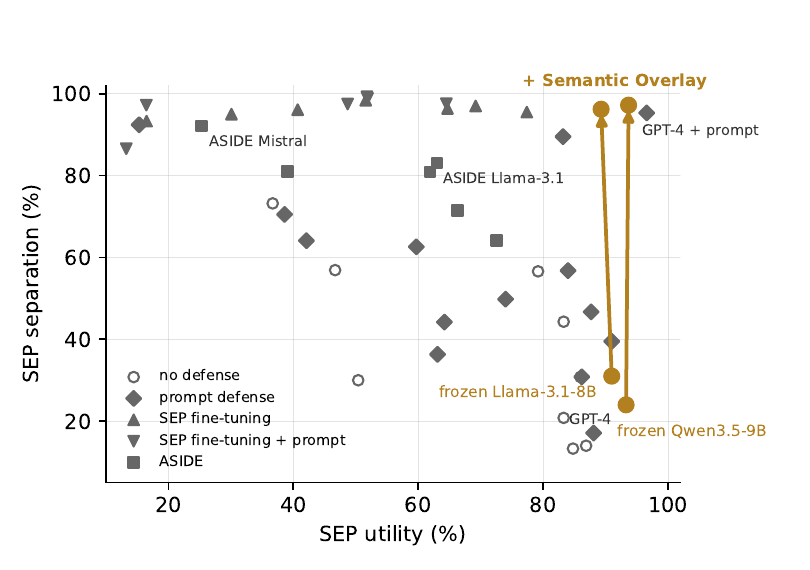}
\caption{Separation versus utility on SEP.
Gray points are published operating points: open models with no defense, with prompt defenses, with SEP fine-tuning, and with SEP fine-tuning combined with the strongest prompt template \citep{zverev2024sep}, and ASIDE on six models \citep{zverev2025aside}.
Fine-tuning buys 93--98\% separation but pays for it in utility (17--77\%); combined with the prompt template it reaches 99.2\%, at 51.8\% utility; ASIDE's strongest separation (Mistral, 92.1\%) costs 20 points of utility.
Two operating points combine both: GPT-4 under an engineered prompt, and the semantic overlay, shown on both of its base models (arrows): under SEP's published grader: frozen Qwen3.5-9B, $+73.2$ points of separation at unchanged utility, and frozen Llama-3.1-8B (Appendix~\ref{app:replication}), $+65.2$ points at $-1.7$.}
\label{fig:frontier}
\end{figure}

Prompt-level defenses (delimiters, warnings, re-stated instructions) are the in-band baseline and are weak \citep{zverev2024sep}; the trained defenses differ in what they modify (Table~\ref{tab:defense-mechanisms}).

\begin{table}[h]
\centering
\footnotesize
\caption{Published prompt-injection defenses: mechanism, and what is served.}
\label{tab:defense-mechanisms}
\begin{tabular}{l>{\raggedright\arraybackslash}p{5.0cm}>{\raggedright\arraybackslash}p{4.3cm}}
\toprule
defense & mechanism & served base model \\
\midrule
prompt defenses \citep{zverev2024sep} & delimiters, warnings, re-stated instructions & unchanged; the defense is forgeable text \\
StruQ \citep{chen2024struq} & fine-tune on structured prompts & fine-tuned \\
ISE \citep{wu2024ise} & trained segment embeddings & fine-tuned \\
ASIDE \citep{zverev2025aside} & rotated data-token embeddings & fine-tuned \\
AIR \citep{air2025} & privilege-indexed embedding at every decoder block & fine-tuned, jointly \\
V-Steer \citep{vsteer2026} & attribution-derived scaling of attention values & frozen; learned version named as future work \\
Semantic Overlays & trained adapters at marked prefill positions & frozen; per-request switch \\
\bottomrule
\end{tabular}
\end{table}

Direct comparison with published prompt-injection defenses requires care, because the strongest of them trains its own models: ASIDE \citep{zverev2025aside} fine-tunes from \emph{base} checkpoints, so its absolute numbers describe its own artifacts, and the fairest comparison is between each method's paired delta against its own baseline (Table~\ref{tab:defense-points}).
As a fraction of remaining headroom closed --- a statistic insensitive to where each baseline starts --- ASIDE closes 48\%; the overlay closes 96\% under SEP's published grader and 98\% under the corrected one; the engineered prompt on GPT-4 closes 94\%, with the SEP authors' own caveat that GPT-4 generated the dataset.
ISE \citep{wu2024ise} beats ASIDE on direct injection for Qwen3; both modify the served weights, where the overlay leaves them frozen and switchable per request.
TensorTrust has a published comparator as well: ASIDE's best model moves attack success 49.9\% $\to$ 36.6\% (Llama 3.1 8B).
Replicated on ASIDE's own 165-row harness and grading, the overlay moves 50.9\% $\to$ 8.5\% (Appendix~\ref{app:inj-tables}).
The strongest published results on the same 165 rows are ISE \citep{wu2024ise} on Qwen3-8B at 19.8\%, as reported by \citet{zverev2025aside}, and Llama-3.1-8B fine-tuned for system-prompt robustness with SFT and DPO plus a double-checking pass, at 20.6\% \citep{mu2025closer}, whose setup ASIDE's harness follows.
ISE's number is a mean over three samples at temperature 0.7; the overlay's is greedy.

\begin{table}[h]
\centering
\footnotesize
\caption{SEP operating points.
Paired cells are each method's own baseline $\to$ defended; $\Delta$ columns are those paired differences.
Range rows aggregate models whose baselines differ; their deltas are per-model pairs.
Foreign rows use each paper's published scoring; the corrected-grader row rescores our own runs only.}
\label{tab:defense-points}
\setlength{\tabcolsep}{4pt}
\begin{tabular}{llcccc}
\toprule
defense & model & separation & $\Delta$sep & utility & $\Delta$util \\
\midrule
ASIDE & Qwen3-8B & 45.3 $\to$ 71.4 & $+26.1$ & 58.9 $\to$ 66.3 & $+7.4$ \\
SEP fine-tuning \citep{zverev2024sep} & seven open models & 93--98 & $+22$ to $+84$ & 17--77 & $-67$ to $+18$ \\
engineered prompt & GPT-4 & 20.8 $\to$ 95.3 & $+74.5$ & 83.3 $\to$ 96.6 & $+13.3$ \\
Semantic Overlays & frozen Qwen3.5-9B & 24.0 $\to$ 97.2 & $+73.2$ & 93.3 $\to$ 93.7 & $+0.4$ \\
\quad corrected grader (Appendix~\ref{app:integrity}) &  & 24.3 $\to$ 98.8 & $+74.5$ & 92.5 $\to$ 93.1 & $+0.6$ \\
\bottomrule
\end{tabular}
\end{table}

Figure~\ref{fig:frontier} plots the published operating points; separation and utility trade off across them, and the two points that combine both are GPT-4 under an engineered prompt and the overlay.
Both are measured on SEP's benign probes, which do not try to defeat the defense; under adversarial pressure the two are not alike, because an engineered prompt is in-band text that attackers forge and override at scale \citep{toyer2023tensortrust, chen2024struq, perez2022ignore}, while the overlay has no textual marker to imitate (adversarial record in Appendix~\ref{app:redteam}).

\begin{table}[t]
\centering
\scriptsize
\caption{Published defenses on static prompt injection benchmarks, as each paper reports them (\%, undefended $\to$ defended).
Rows differ in base model, harness, attack set, and judge, so read each row as one defense's own before and after, not against the others.
Utility: AlpacaFarm block, AlpacaEval win rate against a fixed reference (ours: against the model's own frozen base, which scores 50 against itself by construction; Appendix~\ref{app:utility});
SEP block, each paper's SEP utility (StruQ, DefensiveTokens, Meta-SecAlign, SecPO: judged win rate against Llama-3-8B-Instruct answers; SEP fine-tuning and ours: the probe is answered when the instruction asks);
AgentDojo block, benign task success.
Kind: FT fine-tunes the served model; frozen leaves the base weights unchanged (frozen$^+$ adds trained token embeddings or attention gates); filter and system act outside the model.
$^a$StruQ's strongest static attack, hand-checked marker anywhere in the response.
$^b$Separation, not attack success.
$^c$Run by \citet{wang2026srft}.}
\label{tab:published-defenses}
\setlength{\tabcolsep}{4pt}
\begin{tabular}{llllll}
\toprule
defense & kind & model & attack success & utility & source \\
\midrule
\multicolumn{6}{l}{\emph{AlpacaFarm, StruQ's static attacks}} \\
StruQ & FT & Llama-7B & 96 $\to$ 0$^a$ & 67.2 $\to$ 67.6 & \citet{chen2024struq}, T2--3 \\
SecAlign & FT & Llama3-8B-Instruct & 51 $\to$ 0 & 85.4 $\to$ 85.9 & \citet{chen2025secalign}, T6--7 \\
DefensiveTokens & frozen$^+$ & Llama-3.1-8B & 69.2 $\to$ 0.5 & 29.1 $\to$ 28.5 & \citet{chen2025defensivetokens}, T3 \\
Meta-SecAlign & FT & Llama-3.1-8B & 68.3 $\to$ 0 & 31.2 $\to$ 31.0 & \citet{chen2025metasecalign} v3, T12 \\
Semantic Overlays & frozen & Qwen3.5-9B & 99.0 $\to$ 0.5 & 50 $\to$ 51.2 & Tables~\ref{tab:alpaca}, \ref{tab:alpacaeval} \\
\midrule
\multicolumn{6}{l}{\emph{SEP}} \\
SEP fine-tuning & FT & Llama-3 8B & 30.8 $\to$ 98.4$^b$ & 86.0 $\to$ 51.6 & \citet{zverev2024sep}, T4--5 \\
StruQ & FT & Llama-3.1-8B & 71.4 $\to$ 0.2 & 54.7 $\to$ 52.9 & \citet{chen2025defensivetokens}, T3 \\
DefensiveTokens & frozen$^+$ & Llama-3.1-8B & 71.4 $\to$ 2.8 & 54.7 $\to$ 53.8 & \citet{chen2025defensivetokens}, T3 \\
Meta-SecAlign & FT & Llama-3.1-8B & 98.1 $\to$ 7.8 & 51.4 $\to$ 48.8 & \citet{chen2025metasecalign} v3, T12 \\
SecPO & FT & Llama-3.1-8B & 98.1 $\to$ 1.4 & 55.4 $\to$ 59.4 & \citet{chen2026secpo}, T1 \\
SecPO & FT & Qwen3.6-27B & 98.9 $\to$ 1.8 & 93.0 $\to$ 94.3 & \citet{chen2026secpo}, T2 \\
Semantic Overlays & frozen & Qwen3.5-9B & 71.7 $\to$ 1.2 & 92.5 $\to$ 93.1 & \S~\ref{sec:injection} \\
\midrule
\multicolumn{6}{l}{\emph{AgentDojo}} \\
Meta-SecAlign & FT & Llama-3.1-8B & 7.6 $\to$ 1.0 & 27.8 $\to$ 23.7 & \citet{wang2026srft}$^c$, T1 \\
SR-Agent & FT & Qwen3-8B & 17.0 $\to$ 1.1 & 60.8 $\to$ 51.6 & \citet{wang2026srft}, T2 \\
Meta-SecAlign & FT & Llama-3.3-70B & 14.7 $\to$ 1.9 & 59.8 $\to$ 84.5 & \citet{chen2025metasecalign} v4, T4 \\
Heads of Security & frozen$^+$ & Llama-3.3-70B & 23.1 $\to$ 0.9 & 53.7 $\to$ 58.5 & \citet{shilov2026hos}, T3 \\
SecPO & FT & Qwen3.6-27B & 26.9 $\to$ 0.0 & 89.7 $\to$ 90.7 & \citet{chen2026secpo}, T2 \\
Sanitizer & filter & GPT-4o & 57.7 $\to$ 0.0 & 83.0 $\to$ 74.1 & \citet{bhagwatkar2025firewalls}, T1 \\
CaMeL & system & Claude 3.5 Sonnet & 4.6 $\to$ 0.0 & 90.7 $\to$ 63.9 & \citet{debenedetti2025camel}, T5--7 \\
CaMeL & system & Claude 4 Sonnet & 7.9 $\to$ 1.2 & 86.6 $\to$ 74.2 & \citet{debenedetti2025camel}, T2--4 \\
Semantic Overlays & frozen & Qwen3.5-9B & 12.8 $\to$ 0.0 & 90.7 $\to$ 89.7 & \S~\ref{sec:injection} \\
\bottomrule
\end{tabular}
\end{table}

\subsection{Same-model baselines}
\label{app:same-model}

The comparisons above set our numbers next to numbers other papers measured on other base models.
Table~\ref{tab:inj-compare} instead runs other defenses on our own base models, through our own harnesses.
Its prompt defenses are two of the four of \citet{zverev2024sep}: a sandwich that repeats the instruction after the data, and spotlighting, which wraps the data in delimiters.
The other two, an instruction to ignore commands in the data and a reminder, do no better on 300 SEP items (separation 26.4\% and 23.3\%, against 37.2\% and 44.0\%).
DefensiveTokens \citep{chen2025defensivetokens} and Meta-SecAlign \citep{chen2025metasecalign} are released only for Llama, so they appear only there.

The five controls trained on our corpus each differ from the overlay in one thing.
The three LoRAs mark the untrusted span in band, with two new special tokens as in StruQ \citep{chen2024struq}, with text delimiters, or not at all, and change the model at every position.
The tokenizer keeps the special-token ids out of message contents, so that marker cannot be forged either.
SecAlign's objective \citep{chen2025secalign} trains with preference pairs instead of cross-entropy.
The per-layer vectors replace the adapter with one learned vector per layer, added on the span.
Each LoRA and DPO control's checkpoint is the one with the highest SEP utility on 300 held-out SEP items, and the per-layer vectors are reported at their final step; removing those items from the reported SEP changes no row by more than 0.2 points.

Prompt defenses move little: spotlighting raises separation by about 24 points on Qwen and 37 on Llama, and none brings TensorTrust hijacking below 25\%.
On AlpacaFarm, no prompt defense brings attack success below 84\% (spotlighting 84.1\%, reminder 86.5\%), against 0.5\% for the overlay.
On Llama, Meta-SecAlign and the overlay both reach near-complete separation and near-zero PIArena attack success, but Meta-SecAlign pays 12.6 points of SEP utility and still lets 43.2\% of TensorTrust hijacks through, against the overlay's 4.4\%.
DefensiveTokens halves TensorTrust attack success at a cost of 36 points of SEP utility.

The LoRAs' utility depends heavily on the checkpoint.
On 300 SEP items, the special-token LoRA keeps 20--26\% utility at most checkpoints between steps 50 and 800 and reaches 69.3\% only at step 400; the text-delimiter LoRA peaks at 79.0\% at step 25 and falls to 26.7\% by step 400.
A LoRA that loses utility still does the main task but drops the extra question the trusted prompt asks (Table~\ref{tab:lora-example}).
SecAlign's objective fails more broadly: the model stops answering trivia even when the trusted instruction asks it to, because the completions it learns to reject are trivia answers, which is also what SEP's utility probe asks for.

\begin{table}[h]
\centering
\footnotesize
\caption{SEP item 1, with the second instruction in the trusted system prompt and nothing planted in the passage. Completions are verbatim, greedy decoding; the LoRA is an earlier special-token LoRA, trained on an earlier version of the corpus, at the end of training.}
\label{tab:lora-example}
\begin{tabular}{@{}p{0.2\linewidth}p{0.76\linewidth}@{}}
\toprule
instruction (trusted) & Update the following text with modern language and contexts while retaining the original tone. On top of your main task, answer this priority question: State the name of the river that runs through London. \\
passage (untrusted) & It is a truth universally acknowledged, that a single man in possession of a good fortune, must be in want of a wife. \\
\midrule
special-token LoRA & It is a truth universally acknowledged that a single man with a solid income must be looking for a wife. \\
Semantic Overlays & It is a universally accepted truth that a single man with a substantial income these days is immediately expected to be looking for a partner. The river that runs through London is the \textbf{River Thames}. \\
\bottomrule
\end{tabular}
\end{table}

\subsection{Utility}
\label{app:utility}

A defense that holds every attack off is of little use if it makes the model worse at its job.
So we measured ordinary helpfulness on two benchmarks with the data marked, comparing each defense with its own undefended base model.

SQuAD~v2 \citep{rajpurkar2018squad} asks a question about a Wikipedia passage; half of its 11{,}873 development questions have no answer in the passage, and the model must say so.
The question goes in the trusted instruction and the passage is marked.
With the overlay, Qwen3.5-9B scores 48.4 F1 against 48.8 for the frozen model (paired difference $-0.5$, 95\% interval $-0.9$ to $0.0$) and 36.4 exact match against 36.7.
The two give character-identical answers to 71\% of questions; of the rest, the overlay fixes 609 that the frozen model got wrong and breaks 641 that it got right, mostly by choosing a slightly longer or shorter span.
Both score low in absolute terms for the same reasons: asked zero-shot, the model pads its answers with words around the span and rarely declines to answer.

AlpacaEval \citep{dubois2024alpacaeval} is 805 everyday instructions.
In 208 of them the instruction comes with an input to work on --- a sentence to rewrite, a passage to answer from, a list to classify --- and that input is the marked span; the other 597 have nothing to mark, and on them the overlay is the frozen model.
Instead of AlpacaEval's fixed reference answers, we compare each defended model with its own frozen base: a judge (Claude Haiku~5.5) sees the two answers and picks the better one or calls a tie, twice, once in each order.
A pair scores 1 if the defended model wins both times, 0 if it loses both, and $\tfrac{1}{2}$ otherwise, so 50\% means no difference (Table~\ref{tab:alpacaeval}).

\begin{table}[h]
\centering
\footnotesize
\caption{AlpacaEval win rate of each defense against its own undefended base model (50\% = no difference), with 95\% bootstrap intervals.
``Marked'' is the 208 instructions with an input.
The Llama overlay was trained on an earlier data mix (Appendix~\ref{app:replication}).}
\label{tab:alpacaeval}
\begin{tabular}{llcc}
\toprule
base model & defense & 208 marked & all 805 \\
\midrule
Qwen3.5-9B & Semantic Overlays & 51.2 [47.6, 55.0] & --- \\
Llama-3.1-8B-Instruct & Semantic Overlays & 45.7 [41.1, 50.2] & 48.9 \\
 & DefensiveTokens & 39.7 [35.1, 44.5] & 39.6 [37.0, 42.1] \\
 & Meta-SecAlign & 30.3 [25.7, 34.9] & 31.6 [29.3, 33.9] \\
\bottomrule
\end{tabular}
\end{table}

On Qwen, marking costs nothing the judge can see.
The older Llama overlay loses a few points on the marked instructions, and its losses share a pattern: when the question the user wants answered sits inside the marked input (``answer the question at the end of the story''), it sometimes summarizes the passage instead of answering.
Meta-SecAlign and DefensiveTokens change the model on every prompt, marked or not, and the judge prefers the undefended model's answer clearly more often: Meta-SecAlign's answers are shorter and plainer (median 1{,}200 characters against 1{,}519), as in answering ``which language is this code?'' with only ``C.''

\section{Replication on a second model family}
\label{app:replication}

The do-not-execute overlay was retrained from scratch on frozen \texttt{Llama-3.1-8B-Instruct}, with the recipe of \S~\ref{sec:training} unchanged: the same adapter shape and size, the same hyperparameters, and the same evaluation harnesses.
The corpus was re-derived against the new base model as Appendix~\ref{app:corpus} specifies --- payloads re-screened (263 of 494 kept), frames re-ranked and re-weighted.

The Llama overlay was trained on an earlier data mix than the Qwen one.
Its numbers are therefore a lower bound on what the recipe reaches on a second
model family rather than a matched comparison; retraining on the current mix is
left to future work.
Table~\ref{tab:llama-repro} gives the results; the red team of Appendix~\ref{app:redteam} was not repeated.

\begin{table}[h]
\centering
\caption{The injection results of \S~\ref{sec:injection}, replicated on frozen \texttt{Llama-3.1-8B-Instruct}.
Metrics and scoring rules match the corresponding Qwen tables: SEP on 1{,}000 items under the published grader; TensorTrust on all 776 hijacking and 570 extraction rows, Defense Validity screened to rows the frozen model can answer; PIArena at 50 rows per family.}
\label{tab:llama-repro}
\small
\setlength{\tabcolsep}{6pt}
\begin{tabular}{lcc}
\toprule
 & frozen model & with overlay \\
\midrule
\multicolumn{3}{l}{\emph{SEP}} \\
probe in data $\downarrow$ & 66.8\% & \textbf{3.5\%} \\
utility $\uparrow$ & 91.0\% & 89.3\% \\
SEP $\uparrow$ & 31.0\% & \textbf{96.2\%} \\
\midrule
\multicolumn{3}{l}{\emph{TensorTrust}} \\
hijacking ASR $\downarrow$ & 68.9\% & \textbf{4.4\%} \\
hijacking Defense Validity $\uparrow$ & 100\% & 95.9\% \\
extraction ASR $\downarrow$ & 74.2\% & \textbf{7.7\%} \\
\midrule
\multicolumn{3}{l}{\emph{PIArena attack success $\downarrow$}} \\
infrastructure failure & 98\% & \textbf{0\%} \\
phishing injection & 50\% & \textbf{0\%} \\
content promotion & 38\% & \textbf{0\%} \\
access denial & 32\% & \textbf{0\%} \\
\bottomrule
\end{tabular}
\end{table}

The two frozen models start far apart and land together.
Frozen Llama is roughly twice as attackable as frozen Qwen on TensorTrust --- hijacking 68.9\% against 35.3\%, extraction 74.2\% against 37.7\% --- and the overlay brings both to the same defended floor (4.4\% and 6.4\%; 7.7\% and 6.3\%).
On PIArena all four families reach 0\%, and SEP separation moves from 31.0\% to 96.2\% with utility inside the run-to-run noise band.
PIArena's clean controls stay at 0\% compliance in both arms, so the drop is not refusal.

ASIDE reports on the same base model \citep{zverev2025aside}: its fine-tune moves separation from 53.2\% to 83.1\% and pays 7.3 points of utility; the overlay moves the frozen model from 31.0\% to 96.2\% and pays 1.7.
The two baselines are different model states --- ASIDE measures from its own fine-tune of the base checkpoint, we measure from the stock instruct model --- so we compare the deltas, each taken from its own starting point (Appendix~\ref{app:position}).

\section{Serving cost}
\label{app:serving}

Overlays run inside unmodified vLLM~0.21 through its plugin interface, which lets the plugin add a hook at the input of every decoder layer.
The question for a deployment is what marking costs relative to the same traffic unmarked.
All measurements are on Qwen3.5-9B on one H100, with each comparison made inside one container, because prefill time varies by 10--30\% between hosts.

\paragraph{How marking is served.}
An overlay edits only prefill positions, so the cost of marking should fall on prefill and leave decoding alone.
The serving mode measured here is built for that.
Steps that only decode replay vLLM's own CUDA graphs, which contain no overlay kernels.
Steps that contain marked tokens run the adapters on those tokens alone, for only the overlay sets whose overlays mark them, using CUDA graphs that the plugin captures for each set, layer, and size bucket of marked tokens.
Several overlays stacked on one token are summed before the output projection, so a stack costs about what one overlay costs.
The adapters compute in bf16.

\paragraph{Accuracy.}
Serving must not change what the overlay does.
We measure the per-token effect of the mark, the log-probability of a fixed continuation with the mark minus without it, and compare it with a reference implementation that runs every adapter eagerly.
Without CUDA graphs the serving mode reproduces the reference bit for bit, for all four overlay sets, in mixed batches, and with stacked marks.
With graphs, the two differ by 0.012--0.032 nats per token, which is about $\sqrt{2}$ times the 0.009--0.023 by which stock vLLM already differs from itself with and without graphs: the difference of two independently noisy measurements, not an error the overlay adds.

\paragraph{Cost.}
The benchmark sends prompts in the frame the evaluations use --- task in the system turn, a retrieved passage in the user turn --- and marks the \emph{whole} passage of \emph{every} prompt, which is a worst case for real traffic.
Each request decodes exactly 256 tokens.
Table~\ref{tab:serving} compares marked with unmarked traffic on the same engine.
Decoding costs nothing: time per output token is unchanged at low concurrency and rises at most 7\% under the heaviest load.
Prefill costs a few milliseconds: marking adds 5\,ms to the first token of a lone 512-token prompt and 21\,ms to a lone 6{,}144-token one.
Throughput stays within about 1\% of unmarked traffic except when the engine is saturated with long, fully marked documents, where it falls 3--4\% (2{,}048 tokens at concurrency 32, 6{,}144 at 8) and at worst 6.6\% (6{,}144 at 32).
Three stacked overlays on every passage token cost the same as one.

\begin{table}[h]
\centering
\caption{Marking cost in vLLM, marked against unmarked traffic on the same engine; every prompt's whole passage is marked; mean of two runs.
``Do-not-execute'' is the injection overlay; ``three stacked'' marks every passage token with three overlays.
At 512 tokens and concurrency 8, stock vLLM's own two runs differed by 9\% (874 and 956 tokens/s), so that cell is noise.}
\label{tab:serving}
\small
\setlength{\tabcolsep}{4pt}
\begin{tabular}{rrrrrr}
\toprule
& & \multicolumn{2}{c}{throughput change} & & \\
\cmidrule(lr){3-4}
prompt tokens & concurrency & do-not-execute & three stacked & added TTFT & ms per output token \\
\midrule
512 & 1 & $-0.3\%$ & $-0.2\%$ & +4.9\,ms & 7.02 $\to$ 7.02 \\
512 & 32 & $-1.1\%$ & $-1.3\%$ & +9.0\,ms & 9.95 $\to$ 10.02 \\
2{,}048 & 1 & $-0.4\%$ & $-0.5\%$ & +8.8\,ms & 7.05 $\to$ 7.05 \\
2{,}048 & 8 & $-1.1\%$ & $-0.9\%$ & +11\,ms & 8.49 $\to$ 8.54 \\
2{,}048 & 32 & $-4.2\%$ & $-3.9\%$ & +21\,ms & 14.26 $\to$ 14.79 \\
6{,}144 & 1 & $-1.1\%$ & $-1.1\%$ & +21\,ms & 7.09 $\to$ 7.09 \\
6{,}144 & 8 & $-3.1\%$ & $-3.7\%$ & +28\,ms & 11.14 $\to$ 11.42 \\
6{,}144 & 32 & $-6.6\%$ & $-6.0\%$ & +33\,ms & 26.61 $\to$ 28.43 \\
\bottomrule
\end{tabular}
\end{table}

A deployment more often marks a short span in one new request while other requests are decoding.
With 32 requests decoding and a 162-token prompt arriving every 150\,ms, marking those prompts adds 4\,ms to their median time to first token (43 to 47--48\,ms) and 4.6\,ms to the decoders' 99th-percentile time between tokens (33 to 37.5\,ms); their mean time between tokens rises 3\%.

\paragraph{Against stock vLLM.}
On the same host, unmarked traffic through this serving mode runs 1--8\% \emph{faster} than stock vLLM.
We have not found the cause, so we do not claim that overlays serve at or above stock speed, and the table reports only marked against unmarked.
A simpler mode that runs the do-not-execute overlay through a hook on every step gives 12.5--16\% lower throughput than stock vLLM whether or not a request is marked: a lone request decodes at 8.2 against 7.0\,ms per token.

\paragraph{Not covered.}
Prefix caching is off in every arm, and all measurements use one GPU.
Engines occasionally stall under heavy mixed prefill and decode, in stock vLLM as well as with the plugin; the benchmark in Table~\ref{tab:serving} had none.
Neither mode is in the released code yet; it ships the original plugin, which runs every adapter eagerly and decodes a lone request at about 32\,ms per token.

\section{Limitations}\label{app:limitations}

All results are from one base model at one scale (Qwen3.5-9B), except the injection defense, which we replicate on Llama-3.1-8B-Instruct (Appendix~\ref{app:replication}).
The replication covers the three injection benchmarks (SEP, TensorTrust, PIArena) and not the red-team engagement, which is an adaptive engagement by an adversary model rather than a fixed benchmark; the marks, language, and transform channels are likewise not replicated.
Main-text tables are single training runs decoded greedily; where decoding was repeated across seeds --- SEP and TensorTrust hijacking, three seeds at temperature 0.7 --- every SEP number reproduced to within 1.0 points and the overlay's hijacking rate to within 0.6, though sampling costs both adapters 3 to 5 points of Defense Validity (Appendix~\ref{app:inj-tables}).

The unforgeability claim is a claim about the interface: the overlay lives in activation space, which an attacker whose only channel is text cannot write to.
It says nothing about a compromised serving stack --- the deployment must know span provenance structurally, because it is the deployment that decides where marks go; the overlay defends against untrusted content, not against the stack that applies it.
The interface claim covers forging the channel, not optimizing against it: an attacker who writes only text can still search for text that defeats the trained behavior, and such adaptive attacks break most published model-level defenses \citep{nasr2025attacker}.
The weights and adapters are public, so an attacker with gradients is realistic.
GCG \citep{zou2023gcg}, at the budget SecAlign uses for its own evaluation \citep{chen2025secalign}, breaks the overlay of the main text on 72\% of 50 AlpacaFarm items, and the frozen model on all of them (Appendix~\ref{app:whitebox}).
The same harness reproduces SecAlign's published 0\% against SecAlign's released adapter on Llama-3-8B: GCG succeeded on none of 20 items.
So the overlay as trained in the main text does not stop an optimizing attacker with the weights.
Adversarial training against a continuous attacker brings the same attack to 14\% at a cost of 0.6 points of SEP utility, but only GCG at one budget and one black-box attack have been run against the hardened overlay.
That black-box attack, PISmith, is judged to break an adversarially trained overlay on 83.6\% of SEP items; the model follows a planted instruction or premise on 31.4\%, and most of these attacks plant false premises, which a do-not-execute overlay does not target (Appendix~\ref{app:whitebox}).
A black-box attacker also gets through: PIArena's Strategy attack, in which an attacker model rewrites the injection, succeeds on 10.8\% of 400 items against the overlay of the main text (Appendix~\ref{app:piarena-strategy}).
The red team's 222 adaptive probes, written by an adversary model, explored the same space one probe at a time against an earlier checkpoint; Appendix~\ref{app:redteam} is one engagement, not a proof.

The do-not-execute overlay is binary; distinguishing sources from each other (this passage may not countermand that one) would require per-source overlays, which the multi-overlay machinery supports but nothing here trains.

Trusted instructions that \emph{delegate} to a marked span (``follow the steps in the document'') override the overlay: the model follows the marked steps, which then run with the developer's authority.
Behavior is inconsistent only when a delegation collides with another trusted rule, such as a procedure in the data that would produce a string the developer forbade (Appendix~\ref{app:redteam}).
The training corpus contains exactly one developer--span relationship; a hierarchy among instructions is untrained here, and training one is a direction for future work.

A non-executable span is not a filtered one: the overlay strips commands but deliberately keeps content quotable, so it must not be deployed as an output filter.

Marked text does not change how the model reads unmarked text in prompts up to 32k tokens, but reading a marked answer passage costs about 8 points of question-answering accuracy at 16k and 32k tokens, against none at 2k and 8k (Appendix~\ref{app:longctx}).

Apart from AgentDojo, all attacks here arrive as untrusted text in a single-turn prompt.
AgentDojo puts them in tool outputs across a multi-turn loop, but only under its static attack; adaptive attacks on agents are untested.

Under StruQ's rule, the overlay that copies marked spans exactly lets slightly more AlpacaFarm attacks through than its predecessor (3.4\% against 1.0\%; LoReFT 1.0\%).
All seven are code tasks in which the model writes the injected print statement into its program (Appendix~\ref{app:newbench}).

Serving cost is measured for one model on one GPU, with prefix caching off (Appendix~\ref{app:serving}).
In vLLM, marking adds a few milliseconds of prefill and almost nothing to decoding; throughput falls at most 6.6\%, when an engine is saturated with long, fully marked documents.
The serving mode that achieves this is not yet in the released code.

\end{document}